\documentclass[10pt,twocolumn,letterpaper]{article}

\usepackage[pagenumbers]{cvpr} 

\title{HandOS: 3D Hand Reconstruction in One Stage 
}

\definecolor{c1}{HTML}{B2DBB9}
\definecolor{c2}{HTML}{F7A6AC}
\definecolor{c3}{HTML}{B8E5FA}
\definecolor{1st}{HTML}{80FF80}
\definecolor{2nd}{HTML}{CCFFCC}
\definecolor{sky}{HTML}{00B0F0}

\definecolor{cvprblue}{rgb}{0.21,0.49,0.74}
\usepackage[pagebackref,breaklinks,colorlinks,allcolors=cvprblue]{hyperref}
\usepackage{multirow}
\usepackage{xcolor}
\usepackage{colortbl} 
\usepackage{marvosym}
\usepackage{cuted}
\usepackage{pifont}%
\usepackage[accsupp]{axessibility}  

\def\paperID{7701} 
\def\confName{CVPR}
\def\confYear{2025}

\author{
Xingyu Chen\textsuperscript{1}\footnotemark[1] \hspace{0.15in}
Zhuheng Song\textsuperscript{3}\footnotemark[1]  \hspace{0.15in}
Xiaoke Jiang\textsuperscript{2} \hspace{0.15in}
Yaoqing Hu\textsuperscript{1} \hspace{0.15in}
Junzhi Yu\textsuperscript{1\Letter} \hspace{0.15in}
Lei Zhang\textsuperscript{2\Letter} \\
\textsuperscript{1} Department of Advanced Manufacturing and Robotics, College of Engineering, Peking University \\
\textsuperscript{2} International Digital Economy Academy (IDEA Research) \\ 
\textsuperscript{3} University of Chinese Academy of Sciences \\
\url{idea-research.github.io/HandOSweb}
}

\begin{document}
\maketitle
\footnotetext[1]{Equal contribution. \Letter\,Corresponding author. This work was done during Xingyu Chen's academic visit at IDEA Research and while Zhuheng Song was an intern at IDEA Research.}

\begin{strip}
    \centering
    \vspace{-30pt}
    \includegraphics[width=\linewidth]{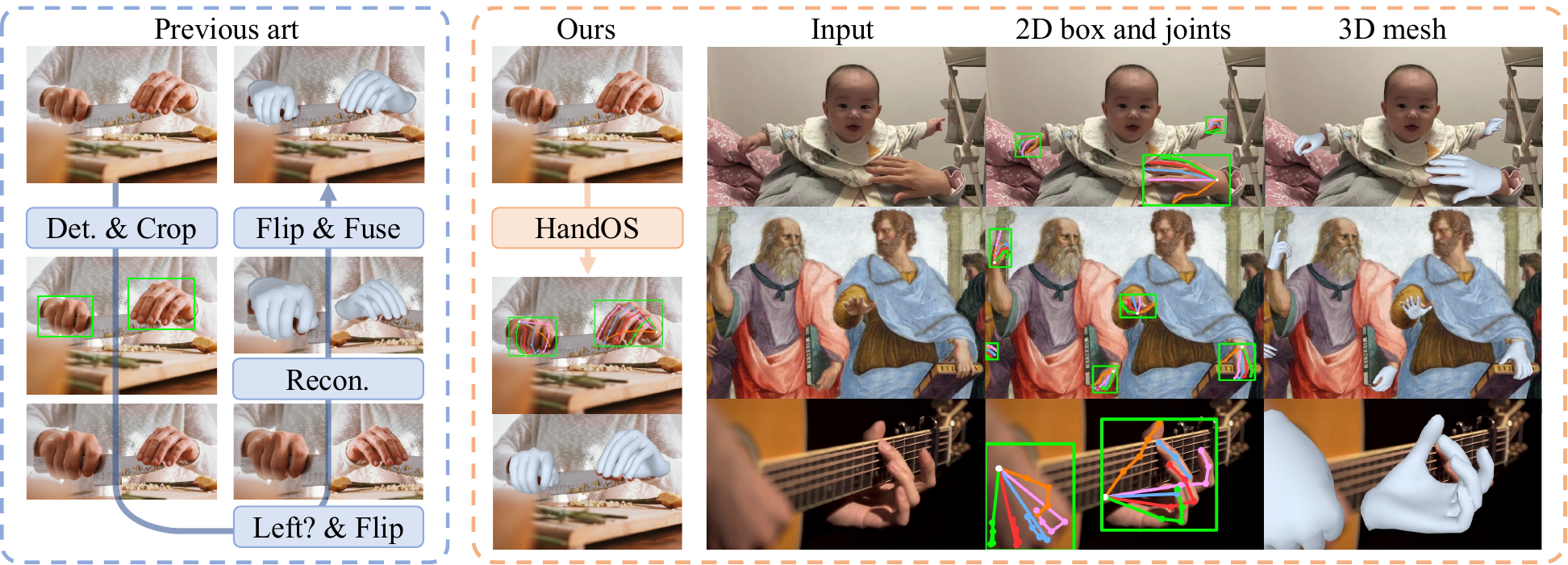}\\
    \captionsetup{type=figure,font=small}
    \caption{We present HandOS, a one-stage approach for hand reconstruction that substantially streamlines the paradigm. Additionally, we demonstrate that HandOS effectively adapts to diverse complex scenarios, making it highly applicable to real-world applications.
    }
    \label{fig:teaser}
\end{strip}

\begin{abstract}
Existing approaches of hand reconstruction predominantly adhere to a multi-stage framework, encompassing detection, left-right classification, and pose estimation. This paradigm induces redundant computation and cumulative errors. In this work, we propose HandOS, an end-to-end framework for 3D hand reconstruction. Our central motivation lies in leveraging a frozen detector as the foundation while incorporating auxiliary modules for 2D and 3D keypoint estimation. In this manner, we integrate the pose estimation capacity into the detection framework, while at the same time obviating the necessity of using the left-right category as a prerequisite. Specifically, we propose an interactive 2D-3D decoder, where 2D joint semantics is derived from detection cues while 3D representation is lifted from those of 2D joints. Furthermore, hierarchical attention is designed to enable the concurrent modeling of 2D joints, 3D vertices, and camera translation. Consequently, we achieve an end-to-end integration of hand detection, 2D pose estimation, and 3D mesh reconstruction within a one-stage framework, so that the above multi-stage drawbacks are overcome. Meanwhile, the HandOS reaches state-of-the-art performances on public benchmarks, e.g., 5.0 PA-MPJPE on FreiHand and 64.6\% PCK@0.05 on HInt-Ego4D.

\end{abstract}    
\section{Introduction}
\label{sec:intro}

The intellectual superiority of humans is expressed through their ability to use the hand to create, shape, and interact with the world. In the era of computer science and intelligence, hand understanding is crucial in reality technique \cite{bib:ARapp,bib:handAR}, behavior understanding \cite{bib:SocialAI,bib:Body2Hands}, interaction modeling \cite{bib:CPF,bib:HopeNet}, embodied intelligence \cite{bib:BiDex,bib:DexGrasp}, and \etc.

Although hand mesh recovery has been studied for years, the pipeline is still confined to a multi-stage paradigm \cite{bib:Baek20,bib:MobRecon,bib:CMR,bib:Metro,bib:Hamba,bib:Hamer}, including detection, left-right recognition, and pose estimation. The necessities behind the multi-stage design are twofold. First, the hand typically occupies a limited resolution within an image, making the extraction of hand pose features from the entire image a formidable task. Hence, the detector is imperative to localize and upscale the hand regions. Second, the pose representation for the left and right hands exhibits symmetry rather than homogeneity \cite{bib:MANO}. Thus, a left-right recognizer is essential to flip the left hand to the right for a uniform pose representation. However, the multi-stage pipeline is computationally redundant, and the performance of pose estimation could be compromised by the dependencies on preceding results. For example, the error rate of detection and left-right classification reaches 11.2\%, when testing ViTPose \cite{bib:vitpose} on HInt test benchmark \cite{bib:Hamer}. That is, some samples are determined to be incapable of yielding accurate results even before the pose estimation process. Therefore, we are inspired to overcome the above challenges by studying an end-to-end framework.

In this paper, we introduce a one-stage hand reconstruction model, termed HandOS, driven by two primary motivations. First, we utilize a pre-trained detector as the foundational model to derive the capacity of 3D reconstruction, with its parameters kept frozen during training. We choose to freeze the detector rather than simultaneously train the detection task because the approach to object detection is already well-studied, and this manner can facilitate data collection while also accelerating convergence. Moreover, to adapt the detector for our tasks, we employ a side-tuning strategy to generate adaptation features.

Second, we adopt a unified keypoint representation (\ie, 2D joints and 3D vertices) for both left and right hands, instead of MANO parameters. It is known that the hand usually occupies a small portion of an entire image, so we design an instance-to-joint query expansion to extract the semantics of 2D joints from the full image guided by detection results. Then, a question naturally arises -- \textit{How to induce 3D semantics with 2D cues and perform 2D-3D information exchange?} To this end, a 2D-to-3D query lifting is proposed to transform 2D queries into 3D space. Besides, considering the different properties between 2D and 3D elements,  hierarchical attention is proposed for efficient training across 2D and 3D domains. Consequently, an interactive 2D-3D decoder is formed, capable of simultaneously modeling 2D joints, 3D vertices, and camera translation.

The contribution of this work lies in three-fold. (1) First of all, we propose an end-to-end HandOS framework for 3D hand reconstruction, where pose estimation is integrated into a frozen detector. Our one-stage superiority is also demonstrated by eliminating the need for prior classification of left and right hands. Therefore, the HandOS framework offers a streamlined architecture that is well-suited for practical real-world applications. (2) We propose an interactive 2D-3D decoder with instance-to-joint query expansion, 2D-to-3D query lifting, and hierarchical attention, which allows for concurrent learning of 2D/3D keypoints and camera position. (3) The HandOS achieves superior performance in reconstruction accuracy via comprehensive evaluations and comparisons with state-of-the-art approaches, \ie, 5.0, 8.4, and 5.2 PA-MPJPE on FreiHand \cite{bib:FreiHand}, HO3Dv3 \cite{bib:HO3D}, and DexYCB \cite{bib:dexycb} benchmarks, along with 64.6\% PCK@0.05 on HInt-Ego4D \cite{bib:Hamer} benchmark.

\section{Related Work}
\label{sec:rw}

\paragraph{3D hand reconstruction.}
Hand reconstruction approaches for monocular image can be broadly categorized into three types. The parametric method \cite{bib:Zhang19,bib:Zhou20,bib:Bihand,bib:ObMan,bib:HIU,bib:Boukhayma19,bib:Baek19,bib:Chen21,bib:TravelNet,bib:Cao21,bib:Baek20,bib:Consist,bib:Liu21,bib:CPF,bib:Zhang21ICCV} typically employ MANO \cite{bib:MANO} as the parametric model and predicts the shape/pose coefficients to infer hand mesh. Voxel approaches \cite{bib:Iqbal18,bib:I2L,bib:Semihand,bib:InterHand} utilize a 2.5D heatmap to represent 3D properties. Lastly, the vertex regression approach estimates the positions of vertices in 3D space \cite{bib:Ge19,bib:YoutubeHand,bib:MobRecon,bib:CMR}.

Recently, the transformer technique \cite{bib:transformer} has been employed to enhance the performance \cite{bib:Metro,bib:MeshGraphormer,bib:Hamba,bib:Hamer,bib:DeFormer,bib:EffHand,bib:PointHMR}. Lin \etal \cite{bib:MeshGraphormer} leveraged the transformer to develop a vertex regression framework, where a graph network is merged with the attention mechanism for structural understanding. Pavlakos \etal \cite{bib:Hamer} utilized the transformer in a parametric framework. Thanks to the integration of 2.7M training data, the generalization ability for in-the-wild images has been significantly enhanced. Dong \etal \cite{bib:Hamba} also developed a transformer-like parametric framework with graph-guided Mamba \cite{bib:mamba}, and a bidirectional Scan is proposed for shape-aware modeling. 
In our framework, we also incorporate the transformer architecture and develop an interactive 2D-3D decoder for learning keypoints in both 2D and 3D domains. 

All of the aforementioned methods adhere to a multi-stage paradigm, including detection and left-right recognition. The purpose of detection is to localize the hand and resize the hand region to a fixed resolution. Rather than utilizing an external detector, we directly enhance the pre-trained detector with the capability to perform pose estimation. Besides, our pipeline eliminates the need for resizing hand regions; instead, we employ instance-to-joint query expansion and deformable attention \cite{bib:defdetr} to extract hand pose features effectively. The purpose of left-right recognition is to flip hand regions, standardizing the representation of left and right hands. In contrast, our approach eliminates the need for this stage, showing that left- and right-hand data can be jointly learned using our 2D-to-3D query lifting and hierarchical attention.

\begin{figure*}[t]
\begin{center}
\vspace{-2em}
\includegraphics[width=\linewidth]{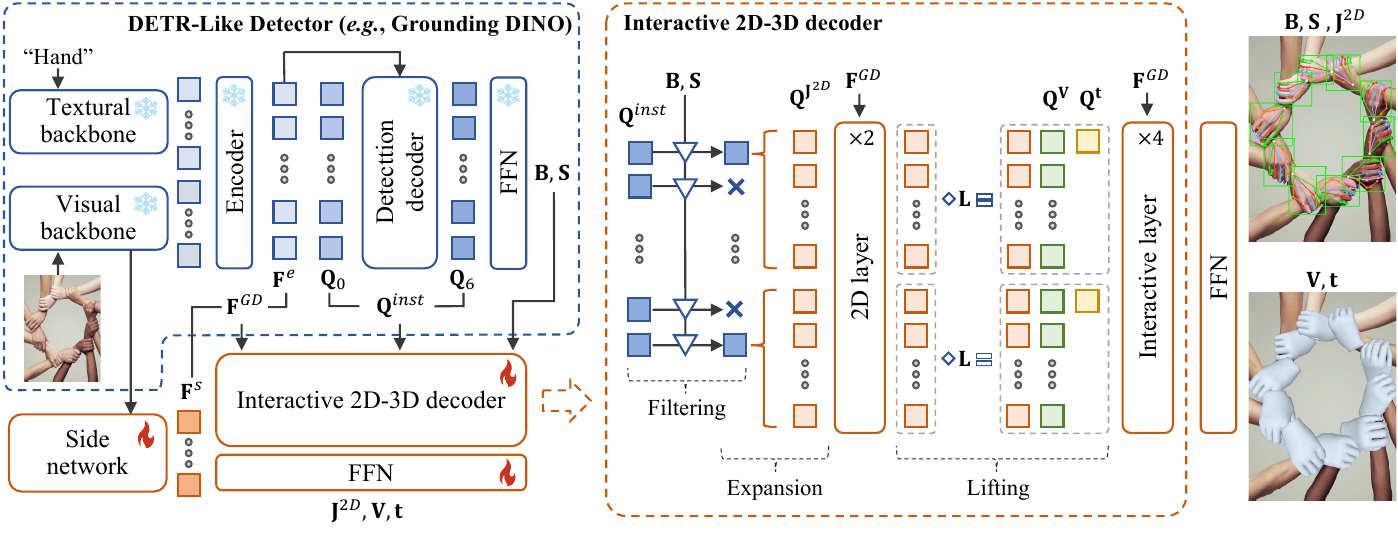}
\caption{
Overview of HandOS framework. Left: overall architecture. Right: interactive 2D-3D decoder. With off-the-shelf features, bounding boxes, and category scores from a frozen detector, the interactive 2D-3D decoder, including query filtering, expansion, lifting, and interactive layers, can understand hand pose and shape via estimating keypoints in both 2D and 3D spaces. Each query $\mathbf Q$ is associated with a reference box, which is not depicted in the figure for conciseness.
}
\label{fig:arch}
\end{center}
\vspace{-0.3cm}
\end{figure*}

\vspace{-0.3cm}
\paragraph{One-stage human pose estimation.}
With the advent of transformer-based object detection \cite{bib:DETR,bib:DINO}, one-stage 2D pose estimation is advancing at a rapid pace. Shi \etal \cite{bib:PETR} proposed PETR, which is the first fully end-to-end pose estimation framework with hierarchical set prediction. Yang \etal \cite{bib:EDPose} designed EDPose with human-to-keypoint decoder and interactive learning strategy to further enhance global and local feature aggregation. 

In the field of whole-body pose estimation, several works have focused on predicting SMPLX \cite{bib:SMPLX} parameters from monocular images in an end-to-end fashion. 
For example, Sun \etal \cite{bib:aios} proposed AiOS, integrating whole-body detection and pose estimation in a coarse-to-fine manner. In contrast, our approach focuses on handling images that contain only hands in a one-stage framework. In addition, instead of utilizing parametric models, we use keypoints to align the representation of left and right hands, while also unifying the representations of 2D and 3D properties.

\vspace{-3mm}
\paragraph{Two-hand reconstruction.}
Although approaches to interaction hands can simultaneously predict the pose of two hands \cite{bib:InterHand,bib:acr,bib:InterWild,bib:MeMaHand,bib:intaghand,bib:decoupled}, they process left and right hands using separate modules and representations. Also, they need to classify the existence of the left and right. In contrast to related works that focus on modeling hand interactions, this paper focuses on single-hand reconstruction with a unified left-right representation.

\section{Method}

Given a single-view image $\mathbf I\in\mathbb R^{H\times W\times 3}$, we aim to infer a 2D joints $\mathbf J^{2D}\in\mathbb R^{J\times 2}$, 3D vertices $\mathbf V\in \mathbb R^{V\times 3}$, and camera translation $\mathbf t\in\mathbb R^{3}$, where $J=21,V=778$. Then, 3D joints can be obtained from vertices, \ie, $\mathbf J^{3D}=\mathcal J\mathbf V$, where $\mathcal J$ is the joint regressor defined by MANO \cite{bib:MANO}. With a fixed camera intrinsics $\mathbf K$, 3D joints can be projected into image space, \ie, $\mathbf J^{proj}=\Pi_\mathbf K(\mathbf J^{3D}+\mathbf t)$, where $\Pi$ is the projection function. The overall framework is shown in \autoref{fig:arch}.

\subsection{Prerequisite: Grounding DINO}
DETR-like detectors can serve as the foundation for HandOS. For instance, Grounding DINO \cite{bib:GD} is utilized, which can detect objects with text prompts. In particular, we use ``Hand'' as the prompt without distinguishing the left and right. Referring to \autoref{fig:arch}, Grounding DINO is a transformer-based architecture with a visual backbone $\mathcal B^v$, a textual backbone $\mathcal B^t$, an encoder $\mathcal E$, a decoder $\mathcal D$, and a detection head $\mathcal H$. The backbone \cite{bib:eva,bib:bert} takes images or texts as the input and produces features:
\begin{equation}
\begin{aligned}
    &\mathcal B^v: \mathbf I \rightarrow \mathbf F^v\in\mathbb R^{L^v\times d^v}, \quad \mathcal B^t: \mathbf T \rightarrow \mathbf F^t\in\mathbb{R}^{L^t\times d^t},
\end{aligned}
\end{equation}
where $\mathbf F^v$ represents a concatenated 4-scale feature with a flattened spatial resolution. $L^v,L^t$ denote the length of the visual/textural tokens, and $d^v,d^t$ are token dimensions.

The encoder fuses and enhances features to generate a multi-modal representation with 6 encoding layers:
\begin{equation}
    \mathcal E: (\mathbf F^v, \mathbf F^t) \rightarrow \mathbf F^e\in\mathbb R^{T^v\times d^v}.
\end{equation}

The decoder contains 6 decoding layers, aiming at extracting features from $\mathbf F^e$ with deformable attention and refining queries $\mathbf Q\in \mathbb R^{Q\times d^q}$ and reference boxes $\mathbf R\in \mathbb R^{Q\times 4}$, where $Q,d^q$ represent the number and dimension of queries. The decoding layer can be formulated as follows,
\begin{equation}
\begin{aligned}
    \mathcal D: (\mathbf Q, \mathbf R, \mathbf F^e) \rightarrow \mathbf Q, \quad \mathbf R=\texttt{FFN}(\mathbf Q)+\mathbf R,
\end{aligned}
\end{equation}
where \texttt{FFN} denotes feed forward network. Finally, the detection head predicts category scores and bounding boxes:
\begin{equation}
    \mathcal H: (\mathbf Q, \mathbf R) \rightarrow (\mathbf S, \mathbf B)\in\mathbb R^{Q\times T^t}\times \mathbb R^{Q\times 4}.
\end{equation}

As a result, we collect the output in each layer, obtaining an encoding feature set $\mathcal F^e=\{\mathbf F^e_i\}_{i=1}^6$, a query set $\mathcal Q=\{\mathbf Q_i\}_{i=0}^6$. The index $0$ indicates the initial elements before the decoder layers. Finally, the detection results are obtained from the last layer of detection head, producing the bounding box $\mathbf{B}$ and the corresponding score $\mathbf{S}$.

\subsection{Side Tuning}
To maintain off-the-shelf detection capability, we freeze all parameters in the detector. However, as the model is fully tamed for the detection task, keypoint-related representations in $\mathbf{F}^e$ remain insufficient. To conquer this difficulty, we design a learnable network with shadow layers of $\mathcal B^v$ as the input, generating complementary features $\mathbf F^s\in\mathbb R^{T^v\times d^v}$. As a result, the Grounding DINO provides features, \ie, $\mathbf F^{GD}=[\mathbf{F}^e_{6}, \mathbf F^s]$, where $[\cdot,\cdot]$ denotes concatenation. Please refer to \textit{suppl. material} for more details.

\subsection{Interactive 2D-3D Decoder}
The input of decoder consists of $\mathbf{F}^{GD}$, $\mathbf{B}$, $\mathbf{S}$, and queries $\mathbf Q^{inst}=[\mathbf Q_0,\mathbf Q_6]$, while its output includes 2D joints $\mathbf{J}^{2D}$, 3D vertices $\mathbf{V}$, and camera translation $\mathbf{t}$.

\vspace{-0.2cm}
\paragraph{Instance query filtering.}
Grounding DINO produces $Q$ instances but only a part of them belongs to the positive. During training, we employ SimOTA assigner \cite{bib:YOLOX} to assign instances to the positive based on ground truth. SimOTA first computes the pair-wise matching degree, which is represented by the cost $\mathbf C$ between the $i$th prediction and the $j$th ground truth (denoted by ``$\star$''):
\begin{equation}
\begin{aligned}
\mathbf C_{i,j}=&-\mathbf S^\star_j\log(\mathbf S_i)-(1-\mathbf S^\star_j)\log(1-\mathbf S_i) \\ 
&-\log(\texttt{IoU}(\mathbf B_i, \mathbf B^\star_j)).
\end{aligned}
\end{equation}
The cost function incorporates both classification error (\ie, binary cross-entropy) and localization error (\ie, Intersection over Union, IoU). Subsequently, an adaptive number $K$ is derived based on IoU, and top-$K$ instances with the lowest cost are selected as the positive samples \cite{bib:OTA}. The selected queries and boxes are denoted as $\mathbf {\tilde Q}^{inst}\in\mathbb R^{K\times d^q}$ and $\mathbf {\tilde B}\in\mathbb R^{K\times 4}$.

In the inference phase, positive queries are identified by selecting those with a score threshold $T^S$ 
and a NMS threshold $T^{NMS}$.

\vspace{-0.2cm}
\paragraph{Instance-to-joint query expansion.}
We expand instance queries for 2D joint estimation. To this end, a learnable embedding $\mathbf e^{\mathbf Q}\in\mathbb R^{J\times d^q}$ is designed, and joint quires are obtained by adding $\mathbf e^{\mathbf Q}$ with $\mathbf Q^{inst}$:
\begin{equation}
\mathbf Q^{\mathbf J^{2D}}\in\mathbb R^{K\times J\times d^q}=\mathbf {\tilde Q}^{inst}+\mathbf e^{\mathbf Q}.
\end{equation}

The reference boxes for 2D joints $\mathbf R^{\mathbf J^{2D}}\in\mathbb R^{K\times J\times 4}$ can also be derived from instances following EDPose \cite{bib:EDPose}:
\begin{equation}
\begin{aligned}
&\mathbf R^{\mathbf J^{2D}}_{c}\in\mathbb R^{K\times J\times 2}=\texttt{FFN}(\mathbf Q^{\mathbf J^{2D}}) + \mathbf{\tilde B}_c, \\
&\mathbf R^{\mathbf J^{2D}}_{s}\in\mathbb R^{K\times J\times 2}= \mathbf{\tilde B}_s\cdot \mathbf e^{\mathbf R}_{2D},
\end{aligned}
\end{equation}
where the subscript $c,s$ represent the center and size of the reference box, and $\mathbf e^{\mathbf R}_{2D}\in \mathbb R^{J \times 2}$ is the learnable embedding for the box size. $\texttt{FFN}$ denotes feed forward network.

\begin{figure}[t]
\begin{center}
\includegraphics[width=\linewidth]{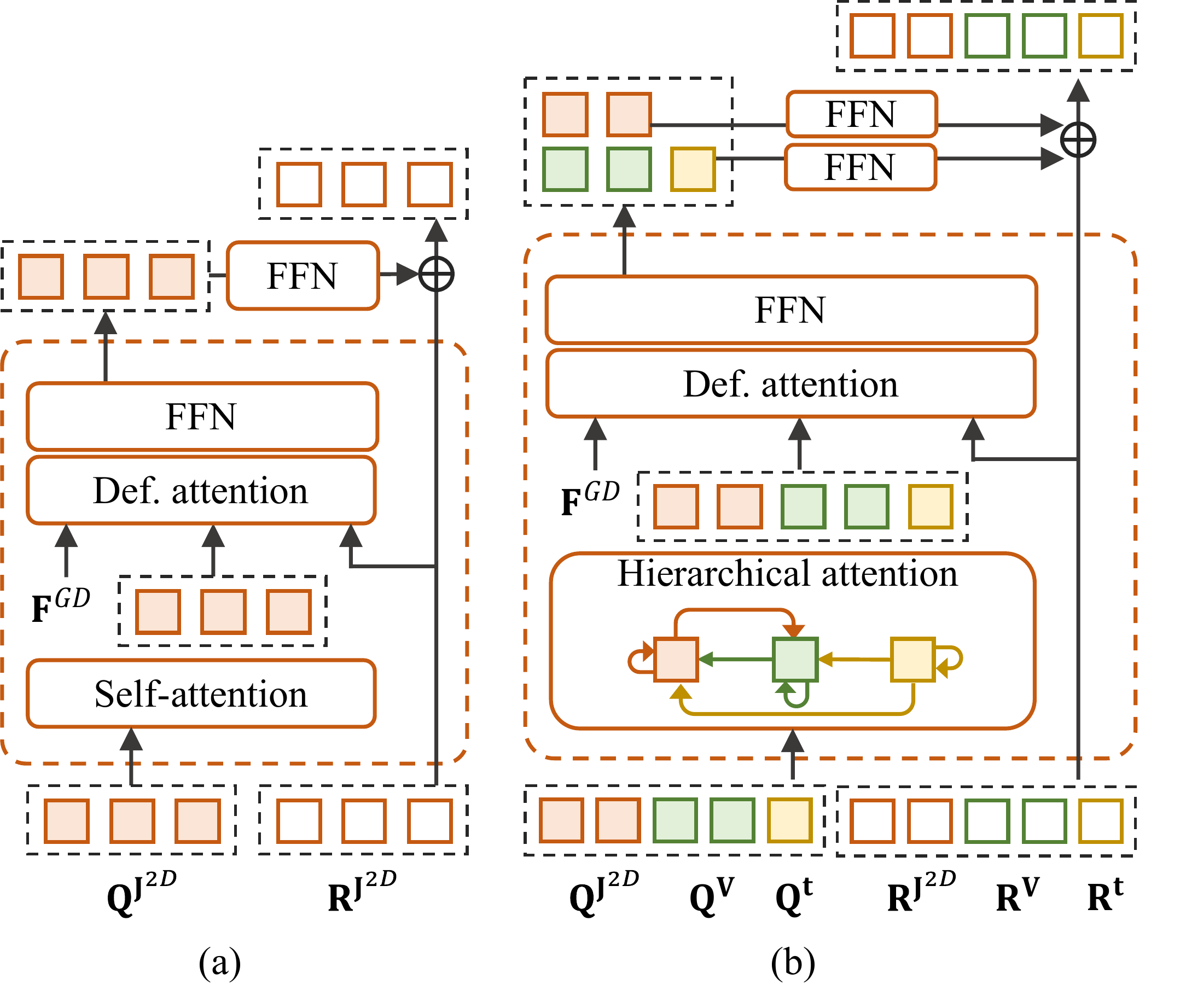}
\vspace{-1.5em}
\caption{
Decoding layers. (a) Canonical 2D layer, popularly employed by previous works. (b) Interactive layer, where hierarchical attention is designed to effectively model 2D and 3D queries. 
}
\label{fig:att}
\vspace{-0.3cm}
\end{center}
\end{figure}

\paragraph{2D-to-3D query lifting.}
Additionally, a third query transformation is employed, namely query lifting, wherein queries and reference boxes are elevated from 2D joints to 3D vertices and camera translation. To this end, we design a learnable lifting matrix $\mathbf L\in\mathbb R^{(V+1)\times J}$ as the weights for the linear combination between 2D and 3D queries, which is initialized with MANO skinning weights. 
Based on $\mathbf L$, the lifting process can be formulated as
\begin{equation}
\begin{aligned}
&[\mathbf Q^{\mathbf V},\mathbf Q^{\mathbf t}]\in\mathbb R^{K\times (V+1)\times d^q}=\mathbf L\diamond \mathbf{Q}^{\mathbf J^{2D}}, \\
&[\mathbf R^{\mathbf V},\mathbf R^{\mathbf t}]\in\mathbb R^{K\times (V+1)\times 4}=\mathbf L\diamond \mathbf{R}^{\mathbf J^{2D}}, \\
&[\mathbf R_c^{\mathbf V},\mathbf R_c^{\mathbf t}]=\texttt{FFN}(\mathbf [\mathbf Q^{\mathbf V},\mathbf Q^{\mathbf t}])+[\mathbf R_c^{\mathbf V},\mathbf R_c^{\mathbf t}], \\
&[\mathbf R_s^{\mathbf V},\mathbf R_s^{\mathbf t}]=[\mathbf R_s^{\mathbf V},\mathbf R_s^{\mathbf t}]\cdot \mathbf e^{\mathbf R}_{3D},
\end{aligned}
\end{equation}
where $\mathbf e^{\mathbf R}_{3D}\in \mathbb R^{(V+1) \times 2}$ is the embedding for the box size, and $\diamond$ denotes Einstein summation for linear combination.

\vspace{-0.2cm}
\paragraph{Decoding layer and hierarchical attention.}
Referring to \autoref{fig:arch}, the decoder comprises 6 layers, with the initial two being designated as 2D layers, and the remaining four functioning as interactive layers. The 2D layer contains self-attention, deformable attention \cite{bib:defdetr}, and FFN in \autoref{fig:att}(a). 

However, the popular design of \autoref{fig:att}(a) cannot directly apply to interactive 2D-3D learning. This is caused by the different properties of 2D joints, 3D vertices, and camera translation: the 2D joints and 3D vertices should exhibit invariance to translation and scale, while the camera parameters should be sensitive for both translation and scale. That is, when the object appears in different positions and scales within the image, the relative structure of the 2D joints $\mathbf J^{2D}$ remains unchanged, the spatial coordinates of the 3D vertices $\mathbf V$ stay constant, while the 3D camera translation $\mathbf t$ varies.
Hence, $\mathbf Q^{\mathbf J^{2D}}$ and $\mathbf Q^{\mathbf V}$ should avoid performing attention operations with $\mathbf Q^{\mathbf t}$. 
Conversely, camera translation is significantly influenced by both 2D position and 3D geometry. Therefore, we enable $\mathbf Q^{\mathbf t}$ to focus its attention on $\mathbf Q^{\mathbf J^{2D}}$ and $\mathbf Q^{\mathbf V}$. Furthermore, 3D vertices provide a robust representation of geometric structure, whereas 2D joints capture rich semantics of image features. Therefore, we allow them to attend to each other, serving as complementary features. 
We refer to this operation as hierarchical attention, through which an interactive layer is formulated as shown in \autoref{fig:att}(b). The arrows in hierarchical attention indicate the visibility of the attention mechanism, with the attention mask as shown in \autoref{fig:att_abl}(c). 

Finally, we use FFN as heads for the regression of 2D joints, 3D vertices, and camera translation.

\subsection{Loss Functions}
\paragraph{2D supervision.}
We use point-wise L1 error and object keypoints similarity (OKS) \cite{bib:yolopose} as the criterion to produce loss terms from 2D annotation (denoted by ``$\star$''):
\begin{equation}
\begin{aligned}
&\mathcal L^{\mathbf J^{2D}}=||\mathbf J^{2D}-\mathbf J^{2D\star}||_1, \\
&\mathcal L^{2D}_{OKS}= \texttt{OKS}(\mathbf J^{2D},\mathbf J^{2D\star}).
\end{aligned}
\end{equation}

\vspace{-0.4cm}
\paragraph{3D supervision.}
We use point-wise L1 error, edge-wise L1 error, and normal similarity  to formulate 3D loss terms:
\begin{equation}
\begin{aligned}
    & \mathcal L^{\mathbf V} = ||\mathbf V-\mathbf V^\star||_1, \quad \mathcal L^{\mathbf J^{3D}} = ||\mathbf J^{3D}-\mathbf J^{3D}||_1, \\
    & \mathcal L^{normal} =\sum\nolimits_{\mathbf{f}\in \mathbb F}\sum\nolimits_{(i,j)\subset\mathbf{f}}|\frac{\mathbf V_i-\mathbf V_j}{||\mathbf V_i-\mathbf V_j||_2}\cdot \mathbf n_\mathbf{f}^\star|, \\
    & \mathcal L^{edge} =\sum_{\mathbf{f}\in \mathbb F}\sum_{(i,j)\subset\mathbf{f}} |||\mathbf V_i-\mathbf V_j||_2 - ||\mathbf V^\star_i-\mathbf V^\star_j||_2|,
\end{aligned}
\end{equation}
where $\mathbb F$ represents mesh faces defined by MANO \cite{bib:MANO}. $\mathcal L^{normal},\mathcal L^{edge}$ are important in our pipeline to induce a rational geometry shape without the aid of MANO inference.

\vspace{-0.4cm}
\paragraph{Weak supervision.}
The majority of samples captured from daily life lack precise 3D annotations. To address this, we introduce weak loss terms based on normal consistency and projection error, enabling the use of 2D annotations for hand mesh learning:
\begin{equation}
\begin{aligned}
&\mathcal L^{\mathbf J^{proj}}=||\mathbf J^{proj}-\mathbf J^{2D\star}||_1, \\
&\mathcal L^{proj}_{OKS}= \texttt{OKS}(\mathbf J^{proj},\mathbf J^{2D\star}), \\
&\mathcal L^{nc}=\sum\nolimits_{\mathbf n_1,\mathbf n_2}(1-<\mathbf n_1, \mathbf n_2>),
\end{aligned}
\end{equation}
where $\mathbf n_1,\mathbf n_2$ are normals of neighboring faces with shared edge, and $<\cdot,\cdot>$ denotes inner product.

Overall, the total loss function is a weighted sum of the above terms, which is applied not only to the final results but also to the intermediate outputs.

\begin{figure}[t]
\begin{center}
\includegraphics[width=\linewidth]{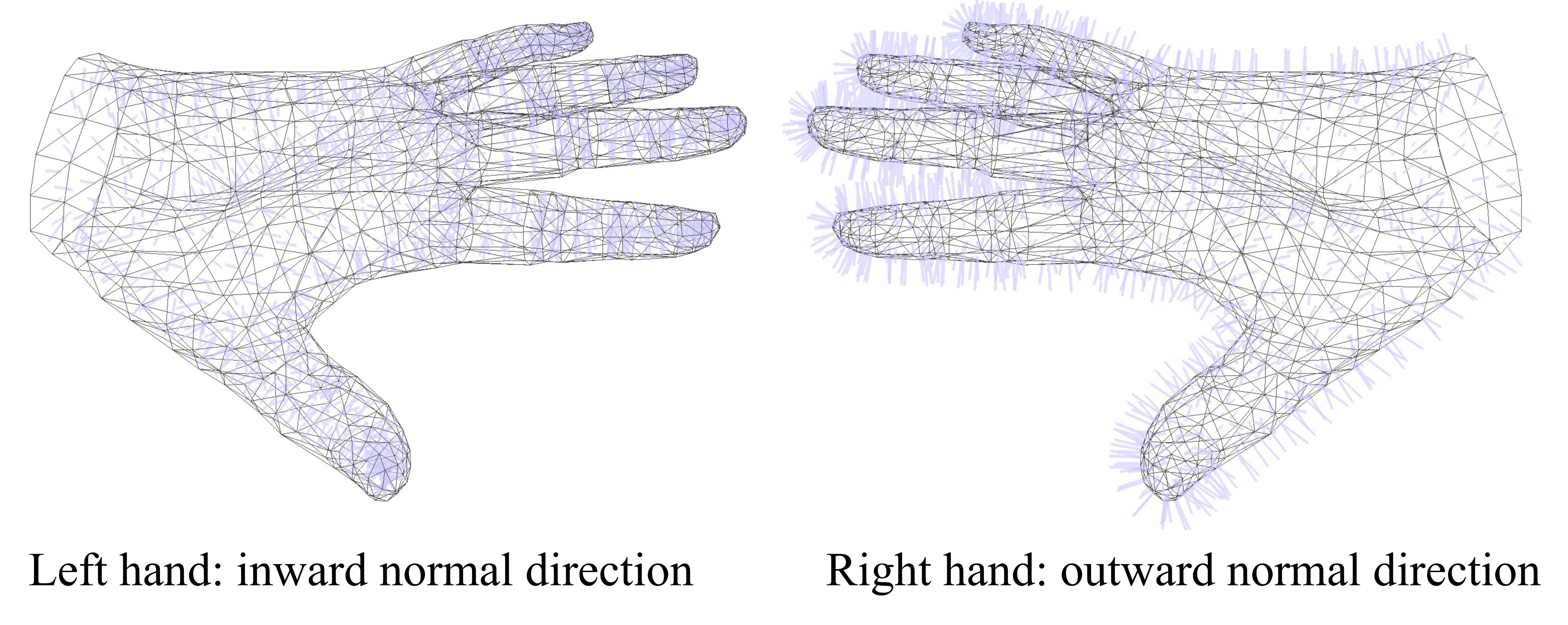}
\vspace{-1.5em}
\caption{Normal vectors serve as left-right indicator. When applying right-hand faces to left or right vertices, the directions of the normal vectors are opposed, as illustrated by the purple lines.
}
\label{fig:normal}
\end{center}
\vspace{-0.3cm}
\end{figure}

\subsection{Normal Vector as Left-Right Indicator}
The HandOS neither requires a left-right category as a prerequisite nor explicitly incorporates a left-right classification module. Nevertheless, the left-right information is already embedded in the reconstructed mesh. Specifically, we use the normal vector as the indicator. As shown in \autoref{fig:normal}, based on the right-hand face, if the mesh belongs to the left hand, the normal vectors point towards the geometric interior; otherwise, they point towards the geometric exterior. In this manner, the left-right category is obtained.

\section{Experiments}

\subsection{Implement Details}

Datasets including FreiHand \cite{bib:FreiHand}, HO3Dv3 \cite{bib:HO3D}, DexYCB \cite{bib:dexycb}, HInt \cite{bib:Hamer}, COCO-WholeBody \cite{bib:cocow}, and Onehand10K \cite{bib:1h10k} are employed for experiments. For the FreiHand, HO3Dv3, and DexYCB benchmarks, we utilize their respective training datasets. To evaluate the HInt benchmark, we aggregate the FreiHand, HInt, COCO-WholeBody, and Onehand10K datasets for training. This combined dataset provides 204K samples, forming a subset of the 2,749K training samples used by HaMeR \cite{bib:Hamer}.

We utilize Grounding DINO 1.5 \cite{bib:GD1.5} as the pre-trained detector to exemplify our approach, noting that our framework is adaptable to other DETR-like detectors. The input is a full image, rather than a cropped hand patch, with its long edge resized to 1280 pixels, following the configuration of Grounding DINO 1.5. We employ the Adam optimizer \cite{bib:Adam} to train our model over 40 epochs with a batch size of 16. The learning rate is initialized at 0.001, with a cosine decay applied from the 25th epoch onward. On the FreiHand dataset, model training takes approximately 6 days using 8 A100-80G GPUs.

PA-MPJPE (abbreviated as PJ), PA-MPVPE (abbreviated as PV), F-socre, PCK, and AUC are used as metrics for evaluation \cite{bib:MobRecon,bib:Hamer} with $T^{S}=0.1,T^{NMS}=0.9$. %

\subsection{Main Results}
We use \colorbox{1st}{Green} and \colorbox{2nd}{Light Green} to indicate the \colorbox{1st}{best} and \colorbox{2nd}{second} results. Previous methods assume that detection and left-right category are accurate, only measuring mesh reconstruction error. In contrast, we do not use the perfect assumption, and our detector achieves 0.44 box AP when measuring hand \cite{bib:cocow} on COCO val2017 \cite{bib:coco}. In terms of missed detection, we use $\mathbf V=\mathbf 0$ for 3D metrics and set $0$ for PCK/AUC. Hence, our results reflect mixed errors across detection, left-right awareness, and mesh reconstruction.

\vspace{-0.4cm}
\paragraph{FreiHand.}
\begin{table}[t]
\small
\renewcommand{\arraystretch}{0.95}
\centering
\begin{tabular}{c | c c c c}
\toprule
Method  & PJ $\downarrow$ & PV $\downarrow$ & F@5 $\uparrow$ & F@15 $\uparrow$ \\
\midrule
METRO \cite{bib:Metro}            & 6.7  & 6.8 & 0.717 & 0.981   
\\
MeshGraphormer \cite{bib:MeshGraphormer} & 5.9  & 6.0 & 0.765 &  0.987   
\\
MobRecon \cite{bib:MobRecon} & \cellcolor{2nd} 5.7  & 5.8 & 0.784 & 0.986   \\
PointHMR \cite{bib:PointHMR} & 6.1 & 6.6 & 0.720 & 0.984 \\
Zhou \etal \cite{bib:EffHand} & \cellcolor{2nd} 5.7 & 6.0 & 0.772 & 0.986 \\
HaMeR \cite{bib:Hamer}       &  6.0 & 5.7 & 0.785 & 0.990   \\
Hamba \cite{bib:Hamba} & 5.8 & \cellcolor{2nd} 5.5 & \cellcolor{2nd} 0.798 & \cellcolor{1st}\bf 0.991 \\
\midrule
HandOS (ours)  & \cellcolor{1st}{\bf 5.0} & \cellcolor{1st}{\bf 5.3} & \cellcolor{1st}{\bf 0.812} & \cellcolor{1st}\bf{0.991} \\
\bottomrule
\end{tabular}
\vspace{-0.6em}
\caption{Results on FreiHand. Errors are measured in mm.}
\label{tab:freihand}
\vspace{-0.1cm}
\end{table}
\begin{table}[t]
\small
\renewcommand{\arraystretch}{0.95}
\centering
\begin{tabular}{c | c c c c}
\toprule
Image flip in training  & PJ $\downarrow$ & PV $\downarrow$ & F@5 $\uparrow$ & F@15 $\uparrow$ \\
\midrule

   \ding{55}  & 5.0 & 5.3 & 0.812 & 0.991 \\
\checkmark & 5.3 & 5.6 & 0.799 & 0.989  \\
\bottomrule
\end{tabular}
\vspace{-0.6em}
\caption{Results on FreiHand with left hands in training data.}
\label{tab:left}
\vspace{-0.3cm}
\end{table}

As shown in \autoref{tab:freihand}, the HandOS demonstrates a notable advantage over prior arts in reconstruction accuracy. Since FreiHand contains only right-hand samples, we flip images to generate left-hand samples for training. According to \autoref{tab:left}, we provide a unified left-right representation and support simultaneous learning for both left and right hands, delivering results comparable to those achieved with right-only training.

\vspace{-0.4cm}
\paragraph{HO3Dv3.}
\begin{table}[t]
\small
\renewcommand{\arraystretch}{0.95}
\setlength\tabcolsep{4pt}
\centering
\begin{tabular}{c | c c c c}
\toprule
Method  & PJ $\downarrow$ & PV $\downarrow$ & F@5 $\uparrow$ & F@15 $\uparrow$ \\
\midrule
AMVUR \cite{bib:amvur} & 8.7 & 8.3 & 0.593 & 0.964  \\ 
SPMHand \cite{bib:spmhand} & 8.8 & 8.6 &0.574 & 0.962 \\ 
Hamba\textsuperscript{*} \cite{bib:Hamba} & \cellcolor{2nd} 6.9 & \cellcolor{2nd} 6.8 & \cellcolor{2nd} 0.681 & \cellcolor{2nd} 0.982 \\ 
\midrule
HandOS (ours) & 8.4 & 8.4 & 0.584 & 0.962 \\
HandOS\textsuperscript{*}  (ours) & \cellcolor{1st}\bf 6.8  & \cellcolor{1st}\bf 6.7 & \cellcolor{1st}\bf 0.688 & \cellcolor{1st}\bf 0.983 \\
\bottomrule
\end{tabular}
\vspace{-0.6em}
\caption{Results on HO3Dv3. Errors are measured in mm. \\
\textsuperscript{*} denotes using extra training data. 
}
\label{tab:ho3d}
\vspace{-0.3cm}
\end{table}
\begin{figure}[t]
\begin{center}
\includegraphics[width=\linewidth]{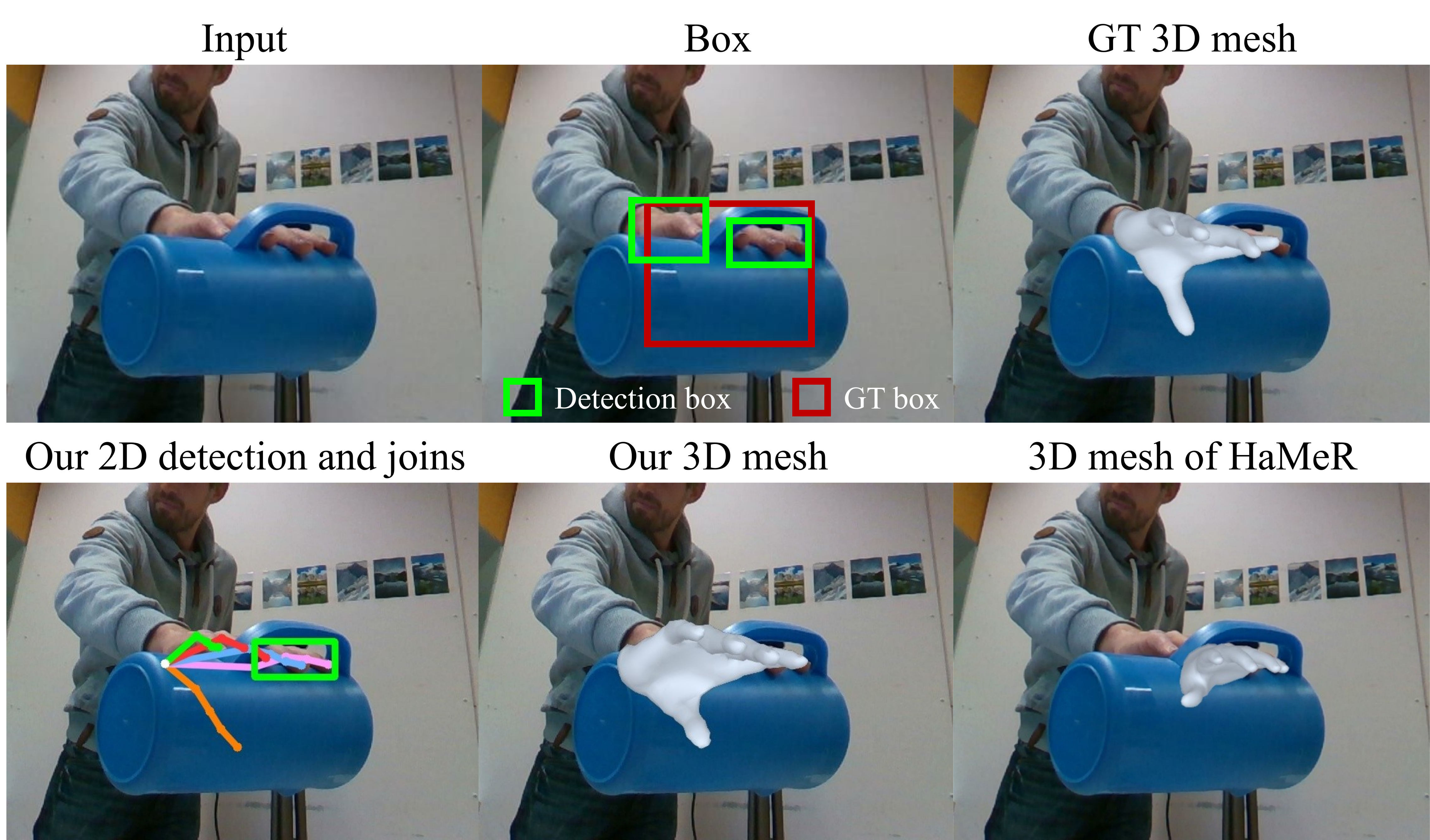}
\vspace{-1.5em}
\caption{
Visualization of HO3Dv3 with actual detection box. We claim that using GT box (red) for downstream tasks is ill-suited. 
}
\label{fig:ho3d}
\end{center}
\vspace{-0.3cm}
\end{figure}
For the scenario of object manipulation, our method also exhibits superior performance, as shown in \autoref{tab:ho3d}. 
However, we claim that the assumption of perfect detection made by precious works is unreasonable for HO3Dv3. Referring to \autoref{fig:ho3d}, for highly occluded sample, only parts of the hand can be detected (\ie, green box), while the ground-truth box still provides a complete hand boundary (\ie, red box) that includes occluded regions. Therefore, the results reported by previous works do not accurately reflect performance in real-world applications.

In contrast, we do not rely on the assumption of perfect detection, and our one-stage pipeline can generate reasonable results from an imperfect box, as shown in \autoref{fig:ho3d}.

\vspace{-0.4cm}
\paragraph{DexYCB.}
\begin{table}[t]
\small
\renewcommand{\arraystretch}{0.95}
\centering
\begin{tabular}{c | c c c}
\toprule
Method  & PJ $\downarrow$ & PV $\downarrow$ & AUC $\uparrow$ \\
\midrule
Spurr \etal. \cite{bib:weakly} & 6.8 & -- & \cellcolor{2nd} 0.864 \\
MobRecon \cite{bib:MobRecon} & 6.4 & 5.6 & -- \\
HandOccNet \cite{bib:HandOcc} & 5.8 & \cellcolor{2nd} 5.5 & -- \\
H2ONet \cite{bib:h2onet} & 5.7 & \cellcolor{2nd} 5.5 & -- \\
Zhou \etal. \cite{bib:EffHand} & \cellcolor{2nd} 5.5 & \cellcolor{2nd} 5.5 & -- \\
\midrule
HandOS (ours) & \cellcolor{1st}\bf 5.2 & \cellcolor{1st}\bf 5.0 & \cellcolor{1st}\bf 0.896  \\
\bottomrule
\end{tabular}
\vspace{-0.6em}
\caption{Results on DexYCB. Errors are measured in mm.}
\label{tab:dex}
\vspace{-0.3cm}
\end{table}
We use DexYCB to further validate the HandOS for object manipulation. 
As shown in \autoref{tab:dex}, we achieve a clear advantage in accuracy over related methods.

\begin{figure*}[t]
\begin{center}
\vspace{-3em}
\includegraphics[width=\linewidth]{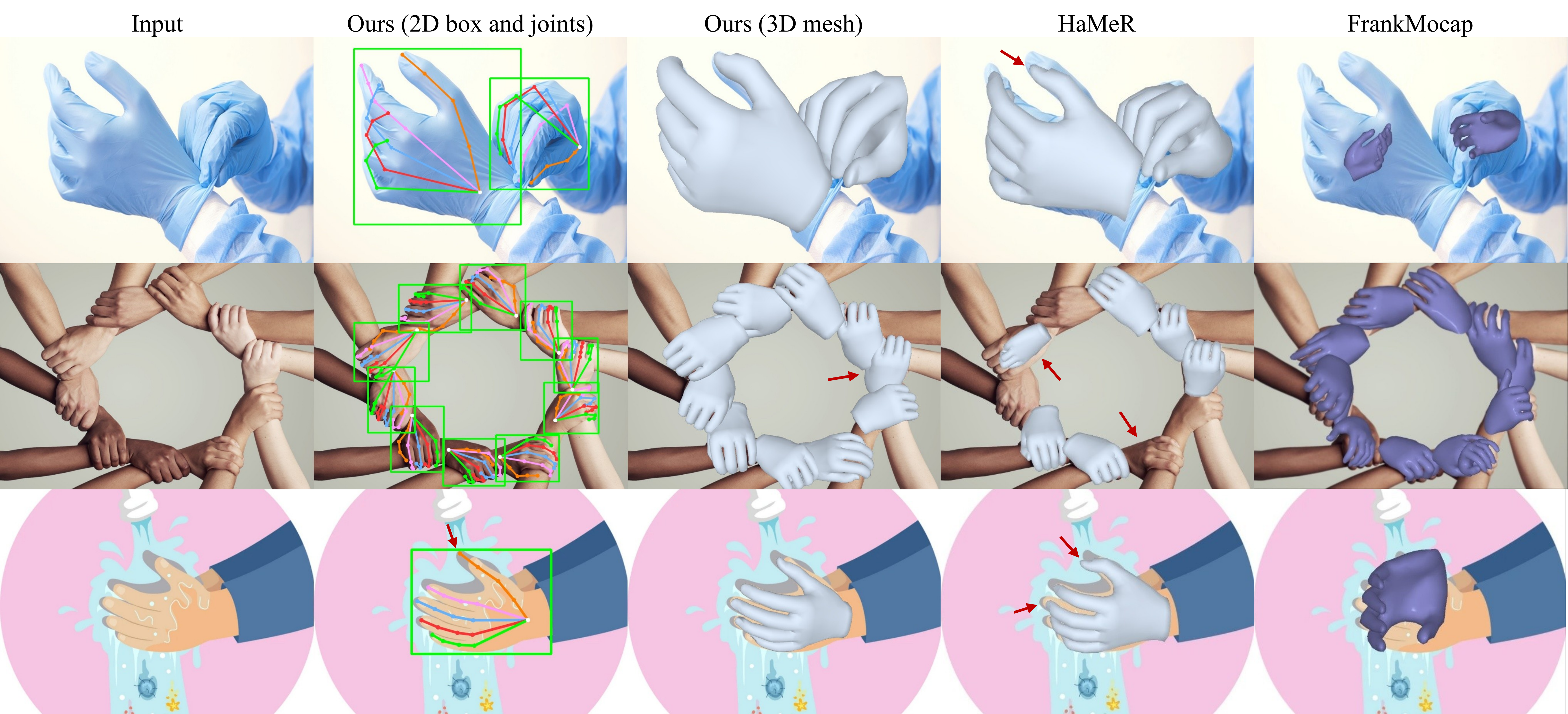}
\vspace{-1.5em}
\caption{
Visual comparison. We are adept at handling long-tail textures, crowded hands, and unseen styles. Red arrows indicate errors.
}
\label{fig:comp}
\end{center}
\vspace{-0.3cm}
\end{figure*}

\vspace{-0.4cm}
\paragraph{Hint.}
\begin{table*}[t]
\small
\renewcommand{\arraystretch}{0.95}
\setlength\tabcolsep{5.5pt}
\centering
\begin{tabular}{c | c c c  c c c  c c c  c c c}
\toprule
& \multirow{2}{*}{Method} & \multirow{2}{*}{Data size} & \multirow{2}{*}{Train Hint} & \multicolumn{3}{c}{New Days}  & \multicolumn{3}{c}{VISOR} & \multicolumn{3}{c}{Ego4D} \\
& & & &  @0.05 & @0.1 & @0.15 & @0.05 & @0.1 & @0.15 & @0.05 & @0.1 & @0.15  \\
\midrule
\multirow{4}{*}{\rotatebox{90}{All}}
& Hamba \cite{bib:Hamba} & 2,720K & & 48.7 & \cellcolor{2nd} 79.2 & \cellcolor{2nd} 90.0 & 47.2 & 80.2 & 91.2 & -- & -- & -- \\
& HaMeR \cite{bib:Hamer} & 2,749K & \checkmark & 51.6 & \cellcolor{1st}\bf 81.9 & \cellcolor{1st}\bf 91.9 & 56.5 & \cellcolor{1st}\bf 88.1 & \cellcolor{1st}\bf 95.6 & 46.9 & 79.3 & 90.4 \\
& HandOS-2D (ours) & \multirow{2}{*}{\bf 204K} & \multirow{2}{*}{\checkmark} & \cellcolor{1st}\bf 55.8 & 75.8 & 84.5 & \cellcolor{1st}\bf 66.2 & 85.3 & 91.8 & \cellcolor{1st}\bf 64.6 & \cellcolor{1st}\bf 85.3 & \cellcolor{2nd} 92.8 \\
& HandOS-proj (ours) & & & \cellcolor{2nd} 53.7 & 75.9 & 85.1 & \cellcolor{2nd} 64.8 & \cellcolor{2nd} 85.4 & \cellcolor{2nd} 92.0 & \cellcolor{2nd} 63.4 & \cellcolor{1st}\bf 85.3 & \cellcolor{1st}\bf 92.9 \\
\midrule
\multirow{4}{*}{\rotatebox{90}{Visible}}
& Hamba \cite{bib:Hamba} & 2,720K & & 61.2 & \cellcolor{2nd} 88.4 & \cellcolor{2nd} 94.9 & 61.4 & 89.6 & 95.6 & -- & -- & -- \\
& HaMeR \cite{bib:Hamer} & 2,749K & \checkmark & 62.9 & \cellcolor{1st}\bf 89.4 & \cellcolor{1st}\bf 95.8 & 66.5 & \cellcolor{1st}\bf 92.7 & \cellcolor{1st}\bf 97.4 & 59.1 & 87.0 & 94.0 \\
& HandOS-2D (ours) & \multirow{2}{*}{\bf 204K} & \multirow{2}{*}{\checkmark} & \cellcolor{1st}\bf 69.8 & 85.0 & 90.6 & \cellcolor{1st}\bf 80.3 & \cellcolor{2nd} 92.5 & \cellcolor{2nd} 95.7 & \cellcolor{1st}\bf 79.7 & \cellcolor{1st}\bf 93.1 & \cellcolor{1st}\bf 96.6 \\
& HandOS-proj (ours) &  & & \cellcolor{2nd} 65.7 & 84.2 & 90.5 & \cellcolor{2nd} 78.1 & 92.1 & 95.6 &  \cellcolor{2nd} 77.4 & \cellcolor{2nd} 92.8 & \cellcolor{1st}\bf 96.6 \\

\midrule
\multirow{4}{*}{\rotatebox{90}{Occluded}}
& Hamba \cite{bib:Hamba} & 2,720K & & 28.2 & 62.8 & \cellcolor{2nd} 81.1 & 29.9 & 66.6 & 84.3 & -- & -- & -- \\
& HaMeR \cite{bib:Hamer} & 2,749K & \checkmark & 33.2 & \cellcolor{1st}\bf 68.4 & \cellcolor{1st}\bf 84.8 & 42.6 & \cellcolor{1st}\bf 79.0 & \cellcolor{1st}\bf 91.3 & 33.1 & 69.8 & 84.9 \\
& HandOS-2D (ours) & \multirow{2}{*}{\bf 204K} & \multirow{2}{*}{\checkmark} & \cellcolor{2nd} 35.5 & 63.4 & 76.1 & \cellcolor{1st}\bf 51.3 & 77.9 & 87.4 & \cellcolor{1st}\bf 46.3 & \cellcolor{2nd} 75.7 & \cellcolor{2nd} 86.9 \\
& HandOS-proj (ours) &  & & \cellcolor{1st}\bf 35.8 & \cellcolor{2nd} 64.4 & 77.5 & \cellcolor{2nd} 50.6 & \cellcolor{2nd} 78.5 & \cellcolor{2nd} 88.0 & \cellcolor{1st}\bf 46.3 & \cellcolor{1st}\bf 76.1 & \cellcolor{1st}\bf 87.3 \\

\bottomrule

\end{tabular}
\vspace{-0.6em}
\caption{Results on HInt. HandOS-2D and HandOS-proj denote the results from our 2D prediction and projected 3D prediction.}
\label{tab:hint}
\vspace{-0.8em}
\end{table*}
We utilize the HInt benchmark with New Days, VISOR, and Ego4D to evaluate HandOS on daily-life images using 2D PCK. In our method, 2D joints can be directly predicted through 2D queries or derived via 3D mesh projection. Accordingly, we report both types of PCK in \autoref{tab:hint}. Compared to prior arts \cite{bib:Hamba,bib:Hamer}, our training dataset is a subset of theirs, with less than one-tenth of their data size. Despite this, HandOS outperforms HaMeR and Hamba across most PCK metrics. Notably, we achieve the highest values on all PCK@0.05 metrics, underscoring the capability for highly accurate predictions. Additionally, we obtain superior values across all metrics on HInt-Ego4D, highlighting our advantage in handling first-person perspectives.

By comparing HandOS-2D and HandOS-proj in \autoref{tab:hint}, it can be concluded that 2D predictions perform better on visible joints, while 3D predictions excel on occluded joints, benefiting from the underlying geometric structure.

\vspace{-0.4cm}
\paragraph{Qualitative results.}
As shown in \autoref{fig:comp},  compared with HaMeR \cite{bib:Hamer} and FrankMocap \cite{bib:Frank},the HandOS can handle complex tasks with long-tail texture, crowded objects, and unseen styles. Even without explicitly classifying left and right hands, we still achieve mostly correct results of left-right awareness and mesh reconstruction in a challenging sample, as shown in the 2nd row. Note that cartoon samples are not involved in our training. Hence, the 3rd row shows our ability to zero-shot generalization across styles. 

Besides, referring to \autoref{fig:ho3d}, HandOS can effectively handle imperfect detection results in occluded scenes. Compared with HaMeR, \autoref{fig:ho3d} can also reflect our one-stage superiority in eliminating cumulative errors. 

\subsection{Ablation Studies}

\paragraph{On pre-trained detector.}
\begin{table}[t]
\small
\renewcommand{\arraystretch}{0.95}
\setlength\tabcolsep{1.5pt}
\centering
\begin{tabular}{c c c c c c | c c c}
\toprule
$\mathbf Q^{inst}$ & $\mathbf Q^{uni}$ & $\mathbf F^e_4$ & $\mathbf F^e_6$ &  $\mathcal B^v$ & SwinT & New Days  & VISOR & Ego4D \\
\midrule
$\checkmark$ &              &    & $\checkmark$ &$\checkmark$ & & \bf 73.8 & \bf 82.5 & \bf 86.7 \\
             & $\checkmark$ &              & $\checkmark$ & $\checkmark$ & &  71.9 & 80.7 & 86.5 \\
$\checkmark$ &              & $\checkmark$  & & $\checkmark$ & & 71.5 & 79.5 & 85.9 \\
$\checkmark$ &              &  & $\checkmark$ &              & $\checkmark$ & 64.7 & 75.6 & 80.1 \\
\bottomrule
\end{tabular}
\vspace{-0.6em}
\caption{Ablation studies on side tuning and feature selection. The number is measured at PCK@0.1. ``$\mathcal B^v$,SwinT'' means that $\mathbf F^s$ is from the pre-trained visual backbone or a from-scratch SwinT.}
\label{tab:side}
\vspace{-0.3cm}
\end{table}
We adapt a pre-trained Grounding DINO for keypoint estimation, making it essential to investigate how the pre-trained model aligns with the downstream task. Key configurations, including query and feature selection as well as side tuning, are examined by a 2D pose model, with their ablation studies detailed in \autoref{tab:side}. 

We use instance queries $\mathbf Q^{inst}\in\mathbb R^{K\times d^q}$ to produce keypoint queries. In another way, keypoint queries can be shared across instances. Hence, we design a unified query $\mathbf Q^{uni}\in\mathbb R^{1\times d^q}$, which is applied to $K$ selected instance with respective reference boxes. In addition, different from $\mathbf Q^{inst}$ that is given by the detector, $\mathbf Q^{uni}$ can be optimized along with the decoder. Referring to the first and second rows of \autoref{tab:side}, $\mathbf Q^{inst}$ has advantages over $\mathbf Q^{uni}$. That is, compared to $\mathbf Q^{uni}$, $\mathbf Q^{inst}$ has instance-specific information that reduces confusion among instances. 

As shown in the first and third rows of \autoref{tab:side}, $\mathbf F^e_6$ outperforms $\mathbf F^e_4$ when used as the value for deformable attention. Since $\mathbf F^e_6$ is the deepest representation, it is significantly influenced by detection training, making it less optimal for keypoint estimation that demands a finer representation of object details. Nevertheless, detailed features can be supplemented through side tuning, enabling $\mathbf F^e_6$ that has the richest semantics to achieve superior performance.

We investigate side tuning by comparing our design with a scratch SwinT network \cite{bib:swin}. As shown in the last rows in \autoref{tab:side}, an additional SwinT trained from scratch induces poor performance. This indicates that, despite being trained on the detection task, the shallow features of $\mathcal V$ can be mapped to adapt to other tasks, with detection pre-training also providing positive benefits.

\vspace{-0.3cm}
\paragraph{Query lifting.}
\begin{figure}[t]
\begin{center}
\includegraphics[width=\linewidth]{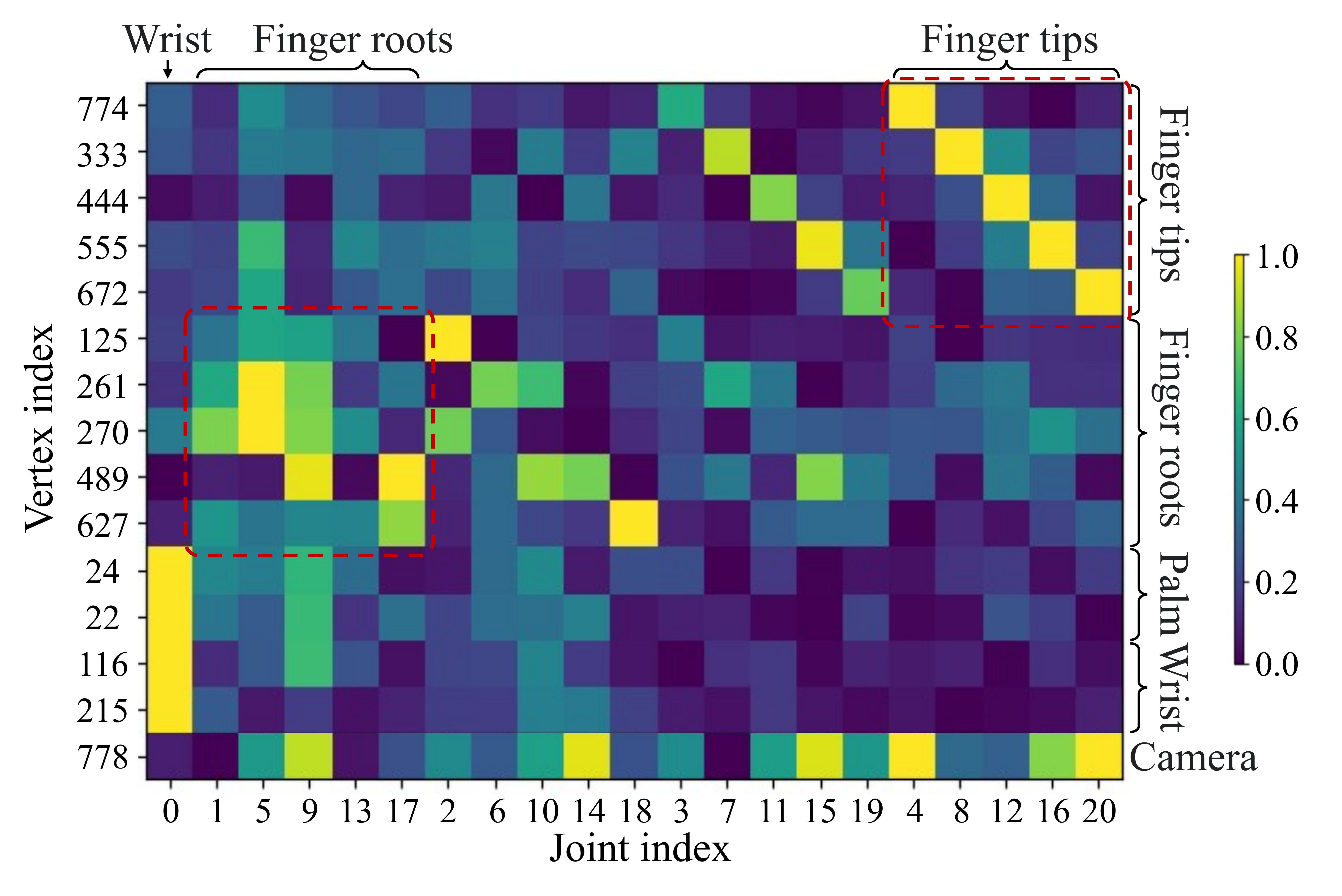}
\vspace{-2em}
\caption{
Query lifting matrix. Tips and roots are arranged in the order of thumb, forefinger, middle finger, ring finger, and pinky. The vertex and joint indices follow MANO and MPII orders.
}
\label{fig:lift}
\end{center}
\vspace{-1em}
\end{figure}
A lifting matrix $\mathbf L$ is designed to transform 2D joint queries to 3D space. In training, $\mathbf{L}$ tends to follow a fixed pattern, and we select several typical lifting patterns for visualization. As illustrated in \autoref{fig:lift}, the lifting process demonstrates semantic consistency, with the vertex queries originating from those of the corresponding joints.

\vspace{-0.3cm}
\paragraph{Hierarchical attention.}
\begin{figure}[t]
\begin{center}
\includegraphics[width=\linewidth]{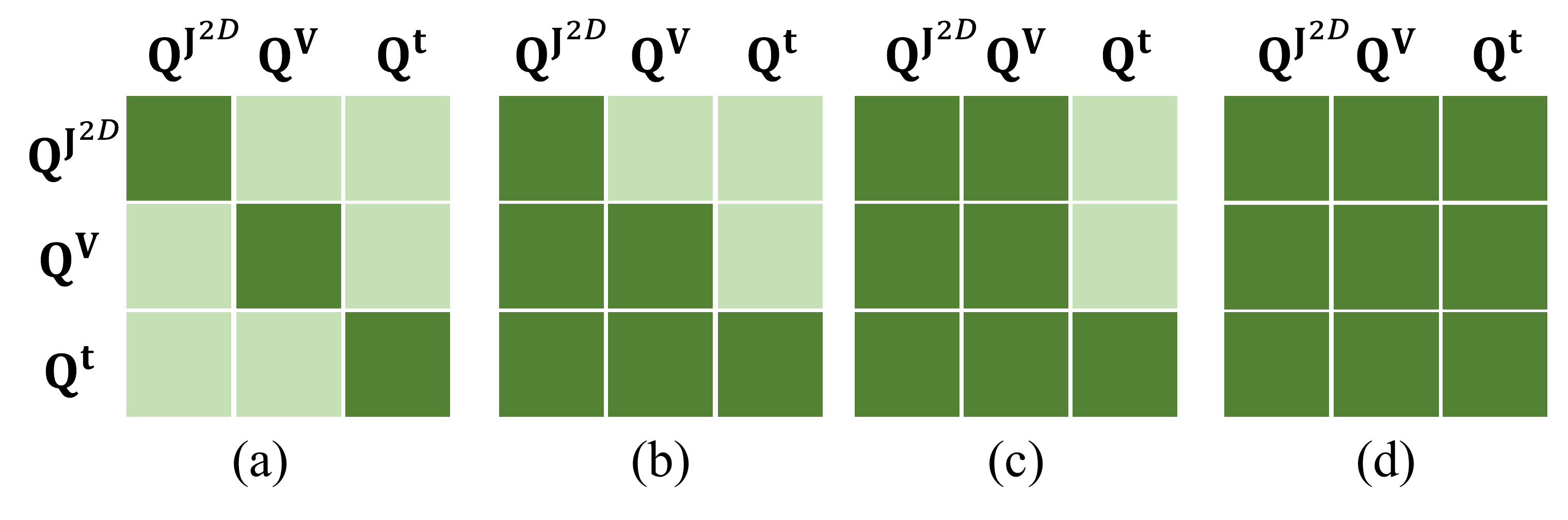}
\vspace{-2em}
\caption{The ablation setting in \autoref{tab:att}. The dark color indicates visible attention computation.
}
\label{fig:att_abl}
\end{center}
\vspace{-1em}
\end{figure}
\begin{table}[t]
\small
\renewcommand{\arraystretch}{0.95}
\setlength\tabcolsep{6pt}
\centering
\begin{tabular}{c | c c c c}
\toprule
 Method & New Days  & VISOR & Ego4D & FreiHand \\
\midrule
\autoref{fig:att_abl}(a) & 75.1/70.9 & 84.9/81.2 & 84.9/82.1 & 6.4 \\
\autoref{fig:att_abl}(b) & 75.7/75.8 & 85.1/85.3 & 85.2/85.2 & 5.7 \\
\autoref{fig:att_abl}(c) & \bf 75.8/75.9 & \bf 85.3/85.4 & \bf 85.3/85.3 & \bf 5.6 \\
\autoref{fig:att_abl}(d) & 74.6/74.8 & 84.1/84.2 & 84.6/84.5 & 5.9 \\

\bottomrule
\end{tabular}
\vspace{-0.6em}
\caption{Ablation studies on attention strategy. The numbers of the Hint benchmark are PCK@0.1 computed with 2D/projected joints. The numbers of FreiHand is PA-MPVPE in mm.}
\label{tab:att}
\vspace{-0.3cm}
\end{table}
Considering the different properties of 2D joints, 3D vertices, and camera translation, we investigate the attention policy and demonstrate the effectiveness of our hierarchical attention, \ie \autoref{fig:att_abl}(c). Compared to it, \autoref{fig:att_abl}(a) produces independent attention across different properties; \autoref{fig:att_abl}(b) is unidirectional attention; and \autoref{fig:att_abl}(d) induces a full attention policy. Referring to \autoref{tab:att}, our design has the best performance in terms of both 2D and 3D metrics. \autoref{fig:att_abl}(a) results in a suboptimal 3D learning due to its invisibility to 2D quires. \autoref{fig:att_abl}(d) make keypoints relevant to camera position, harming the relative space structure of 2D joints and 3D vertices. \autoref{fig:att_abl}(b) has a similar performance to ours, but the interaction among keypoints is insufficient. The necessity of \autoref{fig:att_abl}(c) is also exhibited in \autoref{tab:hint}, where the 2D prediction is good at visible joints, while the projected estimation is adept at occluded joints. This highlights the importance of information exchange between 2D joints and 3D vertices.

\section{Conclusion}
We introduce HandOS, an end-to-end framework for 3D hand mesh reconstruction, which is a unified framework for hand detection, left-right awareness, and pose estimation. Additionally, we propose an interactive 2D-3D decoder with query expansion, lifting, and hierarchical attention, which supports the concurrent learning of 2D joints, 3D vertices, and camera translation. As a result, HandOS achieves state-of-the-art performance on FreiHand, Ho3Dv3, DexYCB, and HInt benchmarks.

%


{
\small
\noindent \textbf{Acknowledgment}
This work was partly supported by the National Natural Science Foundation of China under Grant 62403012, Grant 62233001, Grant U23B2037 and the Postdoctoral Innovative Talent Support Program under Grant BX2023004. The authors acknowledge Ling-Hao Chen for constructive discussions.
}

{
    \small

}

\clearpage

\renewcommand{\thesection}{\Roman{section}}
\renewcommand{\thefigure}{\Roman{figure}}
\renewcommand{\thetable}{\Roman{table}}
\renewcommand{\theequation}{\Roman{equation}}

\def\sectionautorefname{Section}
\begin{abstract}
This is the supplementary document of HandOS, including implementation details (\autoref{sec:imp}), metrics (\autoref{sec:metrics}), discussion on left-right classification (\autoref{sec:cls}), detector adaption (\autoref{sec:det}), and HO3D results (\autoref{sec:ho3d}), as well as more comparison (\autoref{sec:comp}), efficiency analysis (\autoref{sec:flops}), and visual results (\autoref{sec:vis}). Finally, failure cases (\autoref{sec:fail}) and limitations are analyzed (\autoref{sec:limit}).
\end{abstract}

\section{Implementation Details}
\label{sec:imp}

\subsection{Side tuning}

\begin{figure}[h]
\begin{center}
\includegraphics[width=0.9\linewidth]{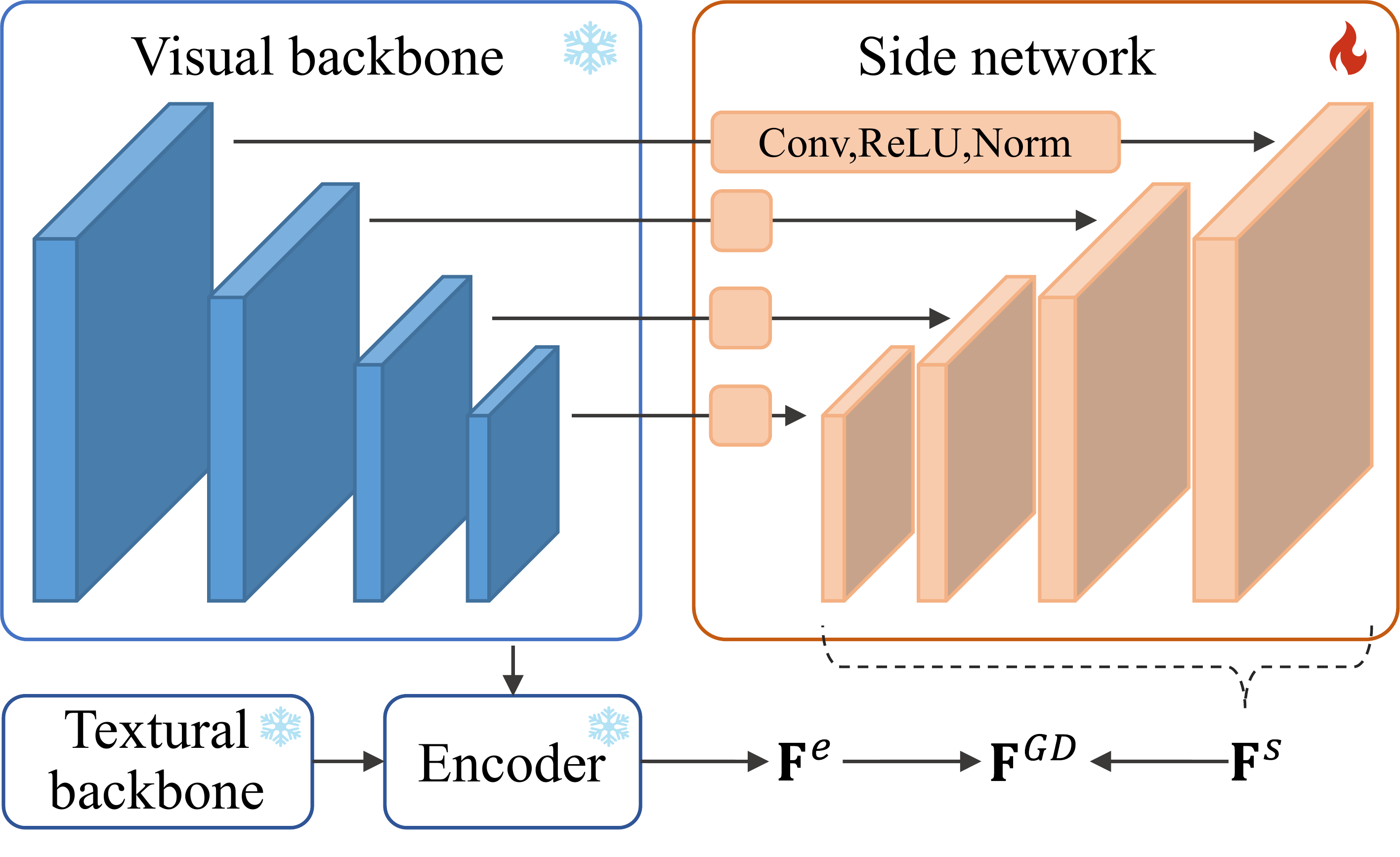}
\vspace{-0.5em}
\caption{
The architecture of side tuning.
}
\label{fig:supp_side}
\vspace{-0.3cm}
\end{center}
\end{figure}

As shown in \autoref{fig:supp_side}, we adopt 4-scale feature maps in the visual backbone. For each scale, we utilize 3 convolution layers for feature mapping. Finally, 4-scale mapped features form $\mathbf F_s$.

\vspace{-0.5cm}
\subsection{Loss function and Training}
The full loss function is given as follows,
\begin{equation}
\begin{aligned}
    \mathcal L &= \lambda^{\mathbf J^{2D}} \mathcal L^{\mathbf J^{2D}} 
    + \lambda^{2D}_{OKS} \mathcal L^{2D}_{OKS}  \\ 
    &+ \lambda^{\mathbf V} \mathcal L^{\mathbf V} 
    + \lambda^{\mathbf J^{3D}} \mathcal L^{\mathbf J^{3D}} \\
    &+ \lambda^{nomral} \mathcal L^{nomral} 
    + \lambda^{edge} \mathcal L^{edge} \\
    &+ \lambda^{\mathbf J^{proj}}  \mathcal L^{\mathbf J^{proj}} 
    + \lambda^{proj}_{OKS} \mathcal L^{proj}_{OKS} \\
    &+ \lambda^{nc} \mathcal L^{nc},
\end{aligned}
\end{equation}
where $\lambda^{\mathbf J^{2D}}=\lambda^{\mathbf J^{3D}}=\lambda^{\mathbf J^\mathbf V}=\lambda^{\mathbf J^{proj}}=\lambda^{edge}=10$, $\lambda^{2D}_{OKS}=\lambda^{proj}_{OKS}=4$, $\lambda^{normal}=5$, $\lambda^{nc}=0.5$.

The HandOS can be trained in an end-to-end manner with $\mathcal L$. To accelerate convergence and reduce experimental time, we adopt a two-stage training. First, a 2D model is trained, whose results are reported in Table~6 of the main text. The 2D model also follows the overall architecture in Fig.~2 of the main text, with all interactive layers replaced by 2D layers. Also, the 2D model does not involve query lifting and 3D vertices/camera prediction. The training data include HInt \cite{bib:Hamer}, COCO \cite{bib:cocow}, and OneHand10K \cite{bib:1h10k}, with the loss function of $\lambda^{\mathbf J^{2D}} \mathcal L^{\mathbf J^{2D}} + \lambda^{2D}_{OKS} \mathcal L^{2D}_{OKS}$. The 2D training cost 3 days on 8 NVIDIA A100 GPUs.

Then, with the weights of the 2D model for initialization, we conduct our experiments on diverse benchmarks with their respective training data.

Ablation studies of loss functions are present in \autoref{tab:loss}. $\mathcal L_{OKS}$ improves the 2D learning efficiency from various-size instances. $\mathcal L^{sp} =\mathcal L^{normal}+\mathcal L^{edge}$ is crucial for structural shape learning, while $\mathcal L^{nc}$ is a smooth regularization. Other losses are strictly required. 

\vspace{-0.25cm}
\begin{table}[h]
\small
\renewcommand{\arraystretch}{1.0}
\centering
\begin{tabular}{c c c | c c}
\toprule
$\mathcal L_{OKS}$  & $\mathcal L^{nc}$ & $\mathcal L^{sp}$ & Ego4D\textsubscript{2D-PCK} & FreiHand\textsubscript{PV} \\

\midrule 
 \checkmark & \checkmark & \checkmark & 85.3 & 5.6 \\
            & \checkmark & \checkmark & 83.2 & 5.8 \\
            &            & \checkmark & 83.2 & 5.9  \\
            &            &            & 82.9 & 13.2   \\
\bottomrule
\end{tabular}
\caption{Ablation study of loss functions.}
\label{tab:loss}
\vspace{-0.3cm}
\end{table}

\section{Metrics}
\label{sec:metrics}

\paragraph{Percentage of correctly localized keypoints (PCK)} is a metric used to evaluate the accuracy of 2D keypoint localization. A keypoint is considered correct if the distance between its predicted and ground truth locations is below a specified threshold.  We use a threshold of 0.05, 0.1, and 0.15 box size, \ie PCK@0.05, PCK@0.1, and PCK@0.15. 

\vspace{-0.3cm}
\paragraph{Mean per joint/vertex position error (MPJPE/MPVPE)} measures the mean per joint/vertex error by Euclidean distance (mm) between the estimated and ground-truth coordinates. Since some global variation cannot be induced from a monocular image, we use Procrustes analysis \cite{bib:PA} to focus on local precision, \ie, PA-MPJPE/MPVPE.

\vspace{-0.3cm}
\paragraph{F-score} represents the harmonic mean of recall and precision calculated between two meshes with respect to a specified distance threshold. Specifically, F@5 and F@15 correspond to thresholds of 5mm and 15mm, respectively.

\vspace{-0.3cm}
\paragraph{Area under the curve (AUC)} represents the area under the PCK curve plotted against error thresholds ranging from 0 to 50mm with 100 steps.

\section{Discussion on Left-Right Classification}
\label{sec:cls}

The recognition of left and right hands is a difficult task. Previous works usually achieve this with body prior \cite{bib:vitpose}. That is, the left and right are easy to understand with whole-body structure. However, there are many scenarios in which the hand appears without a body, such as in egocentric scenes. Here, the classification error increases, harming the performance of the multi-stage method.

Our one-stage pipeline is free from the impact of prior left-right information and uses the normal direction to obtain the left-right category based on the reconstructed mesh. In this manner, as long as the reconstruction results are correct, the left-right hand classification is also accurate. 

Compared with the previous ``left/right $\rightarrow$ mesh'' paradigm, our ``mesh $\rightarrow$ left/right'' investigates another way for hand-side understanding. As a result, our method is superior in left-right classification. Based on the HInt test set, ViTPose \cite{bib:vitpose} achieves a detection recall of 94.6\% and left-right classification precision of 93.8\% with its default settings. In contrast, the HandOS based on Grounding DINO reaches a detection recall of 100\% (with a confidence threshold of 0.1) and left-right classification precision of 97.9\%. Note that the detection precision cannot be calculated since Hint does not label all positive instances in an image.

\section{Adaptation of Other Detector}
\label{sec:det}

We use DINO-X \cite{bib:dinox} as the detector to build the HandOS, which achieve 0.428 box AP when measuring hand category \cite{bib:cocow} on COCO val2017 \cite{bib:coco}. The metrics are shown in \autoref{tab:dinox}, and it is evident that our HandOS is adaptable to all DETR-like detectors.

\vspace{-0.25cm}
\begin{table}[h]
\small
\renewcommand{\arraystretch}{1}
\centering
\begin{tabular}{c | c c c c }
\toprule
Method  &  New Days  & VISOR & Ego4D & FreiHand \\
\midrule 
 main text  & 75.8/75.9 & 85.3/85.4 & 85.3/85.3 & 5.6 \\
 w/ DINO-X  & 76.3/76.5 & 84.8/84.6 & 85.6/85.5 & 5.5 \\
\bottomrule
\end{tabular}
\caption{The numbers of the Hint benchmark are PCK@0.1 computed with 2D/projected joints. The numbers of FreiHand is PA-MPVPE in mm.}
\label{tab:dinox}
\vspace{-0.3cm}
\end{table}

\section{More HO3Dv3 Analysis}
\label{sec:ho3d}

As explained in Fig.~5 of the main text, the inference with the ground-truth box is ill-suited, which is prevalently employed by previous work. We do not follow this setting and use the actual detection box for inference. In addition, the misaligned detection and ground truth could also induce adverse effects for HandOS training, \ie, query filtering based on ground truth becomes less efficient during training. Despite these unfavorable conditions, the HandOS still reaches superior results, \eg 8.4 PA-MPJPE. 

Also, it is necessary to evaluate the model performance with Ho3Dv3 GT boxes. As shown, although GT boxes are not involved in training, the inference can adapt to them, thanks to adaptive within-box feature localization of deformable attention, indicating our robustness to box changes.

To relieve the issue during training, we employ more training data, including FreiHand \cite{bib:FreiHand}, HInt \cite{bib:Hamer}, COCO \cite{bib:cocow}, OneHand10K \cite{bib:1h10k}, HO3Dv3 \cite{bib:HO3D}, DexYCB \cite{bib:dexycb}, CompHand \cite{bib:MobRecon}, and H\textsubscript{2}O3D \cite{bib:kpt_trans}. As shown in \autoref{tab:supp_ho3d}, we achieve state-of-the-art numeric results. Note that our combined training data contains 933K samples, which is smaller than that of Hamba with 2,720K samples.
\begin{table}[h]
\small
\renewcommand{\arraystretch}{1}
\setlength\tabcolsep{4pt}
\centering
\begin{tabular}{c | c c c c}
\toprule
Method  & PJ $\downarrow$ & PV $\downarrow$ & F@5 $\uparrow$ & F@15 $\uparrow$ \\
\midrule
AMVUR \cite{bib:amvur} & 8.7 & 8.3 & 0.593 & 0.964  \\ 

Hamba\textsuperscript{*} \cite{bib:Hamba} & \cellcolor{2nd} 6.9 & \cellcolor{2nd} 6.8 & \cellcolor{2nd} 0.681 & \cellcolor{2nd} 0.982 \\ 
\midrule
HandOS (ours) & 8.4 &  8.4 & 0.584 & 0.962 \\
 w/ GT box (ours) & 8.4 & 8.5 & 0.581 & 0.962 \\
HandOS\textsuperscript{*}  (ours) & \cellcolor{1st}\bf 6.8  & \cellcolor{1st}\bf 6.7 & \cellcolor{1st}\bf 0.688 & \cellcolor{1st}\bf 0.983 \\

\bottomrule
\end{tabular}
\vspace{-0.6em}
\caption{Results on HO3Dv3. Errors are measured in mm. \textsuperscript{*} denotes using extra training data. 
}
\label{tab:supp_ho3d}
\vspace{-0.1cm}
\end{table}

\section{More Qualitative Comparison with HaMeR}
\label{sec:comp}

\begin{figure}[h]
\begin{center}
\includegraphics[width=0.98\linewidth]{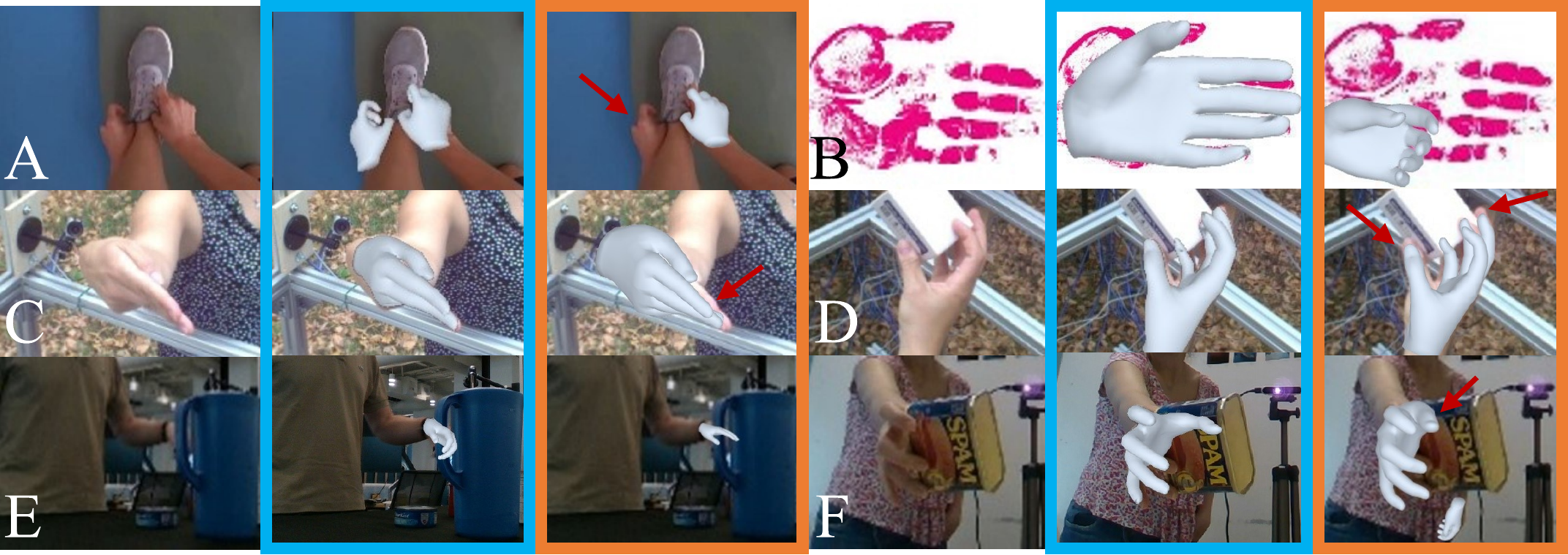}
\caption{Visual comparison between \colorbox{sky}{HandOS} and \colorbox{orange}{HaMeR}.}
\label{fig:rebuttal_comp}
\end{center}
\vspace{-1em}
\end{figure}

More comparisons of HandOS and HaMeR are presented in \autoref{fig:rebuttal_comp}, where we are superior in accurate detection (A), novel-style adaptation (B), fine image alignment with accurate pose/shape (C, D), and reasonable occlusion awareness (E, F).

\section{Comparison of Inference Efficiency.}
\label{sec:flops}

With $P,H$ denoting the number of person and hand, our detector+decoder has (301+108$H$)G FLOPs, using 8G memory; ViTPose+HaMeR has (484$P$+244$H$)G FLOPs, using 12G memory. On RTX3090 and PyTorch, our detector takes 0.5s, and decoder time is from 0.1s ($H$=1) to 0.7s ($H$=10); VitPose+HaMeR takes (0.4+0.06$P$+0.1$H$)s.

\section{Failure Cases}
\label{sec:fail}

\begin{figure}[t]
\begin{center}
\includegraphics[width=\linewidth]{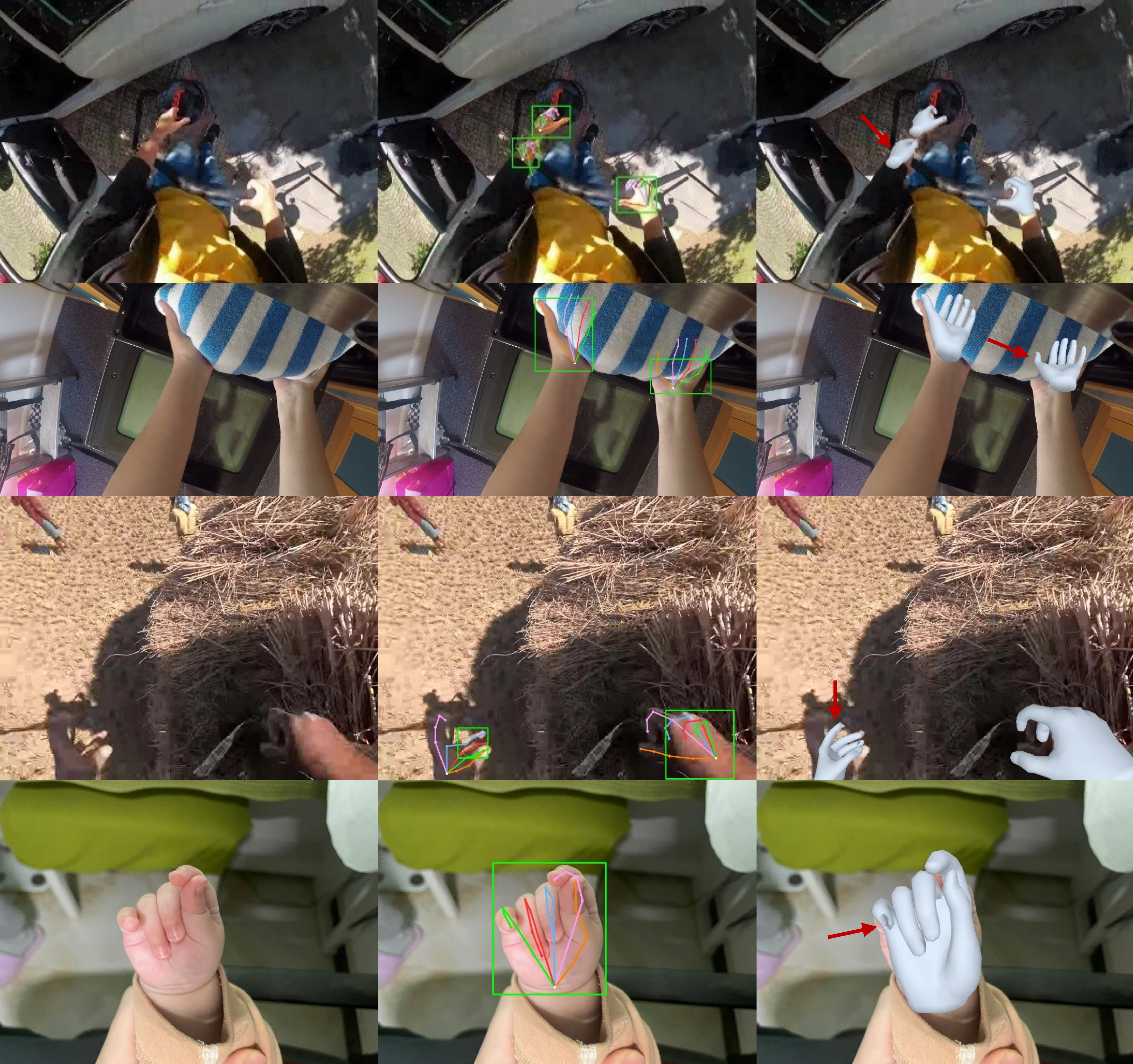}
\vspace{-1.5em}
\caption{
Failure cases. Red arrows indicate errors. Samples in a triplet are input, 2D detection and joints, and 3D mesh.
}
\label{fig:supp_fail}
\vspace{-0.3cm}
\end{center}
\end{figure}

As shown in \autoref{fig:supp_fail}, the HandOS could fail in false positive (the 1st row), left-right awareness (the 2nd row), inaccurate pose (the 3rd row), and geometry artifacts (the 4th row), when handling extreme lighting, occlusion, and shape conditions.

\section{Qualitative Results}
\label{sec:vis}

\label{sec:vis}
\begin{figure*}[t]
\begin{center}
\includegraphics[width=\linewidth]{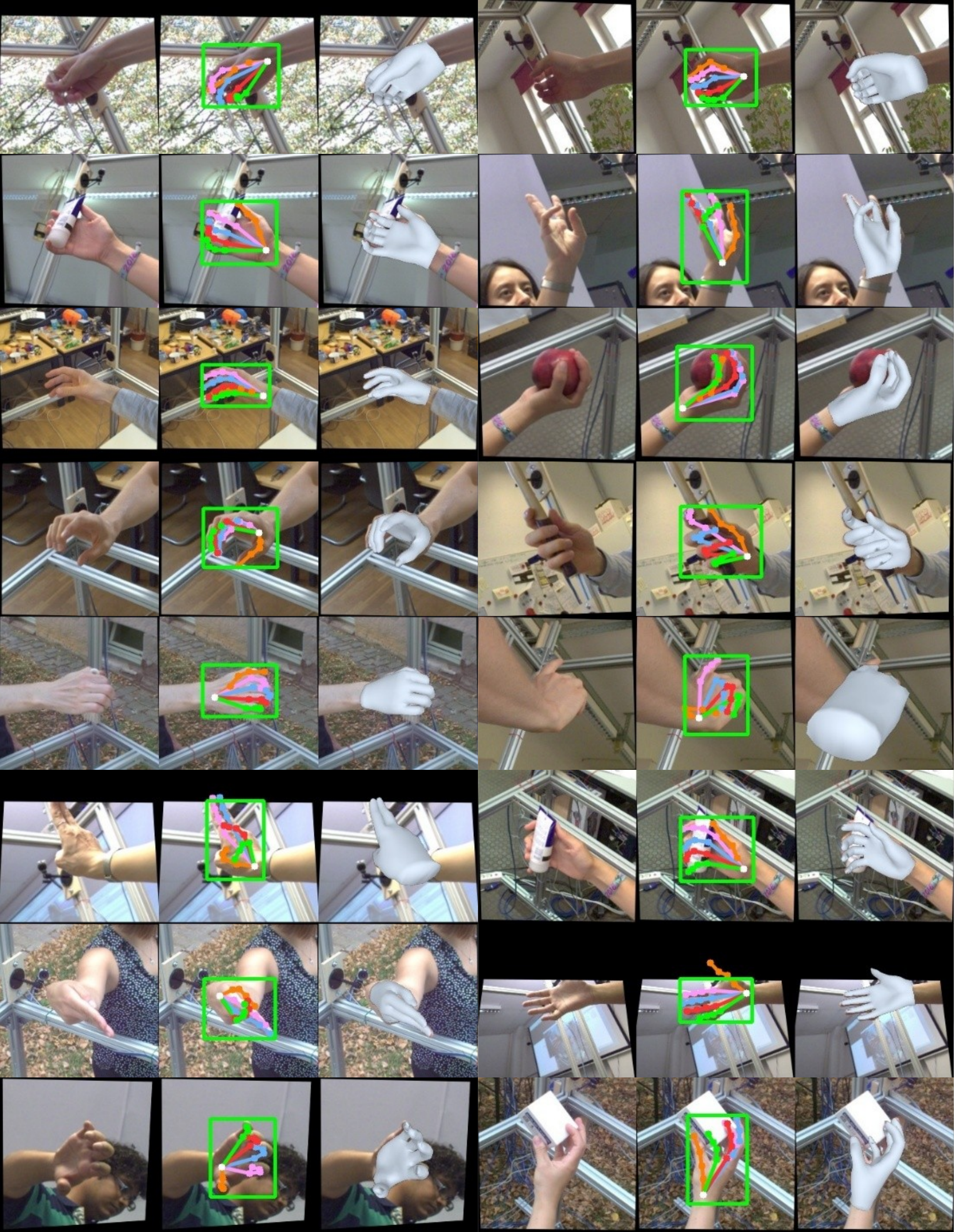}
\caption{
Visualization of FreiHand evaluation set. Samples in a triplet are input, 2D detection and joints, and 3D mesh. 
}
\label{fig:supp_frei}
\end{center}
\vspace{-0.3cm}
\end{figure*}

\begin{figure*}[t]
\begin{center}
\includegraphics[width=\linewidth]{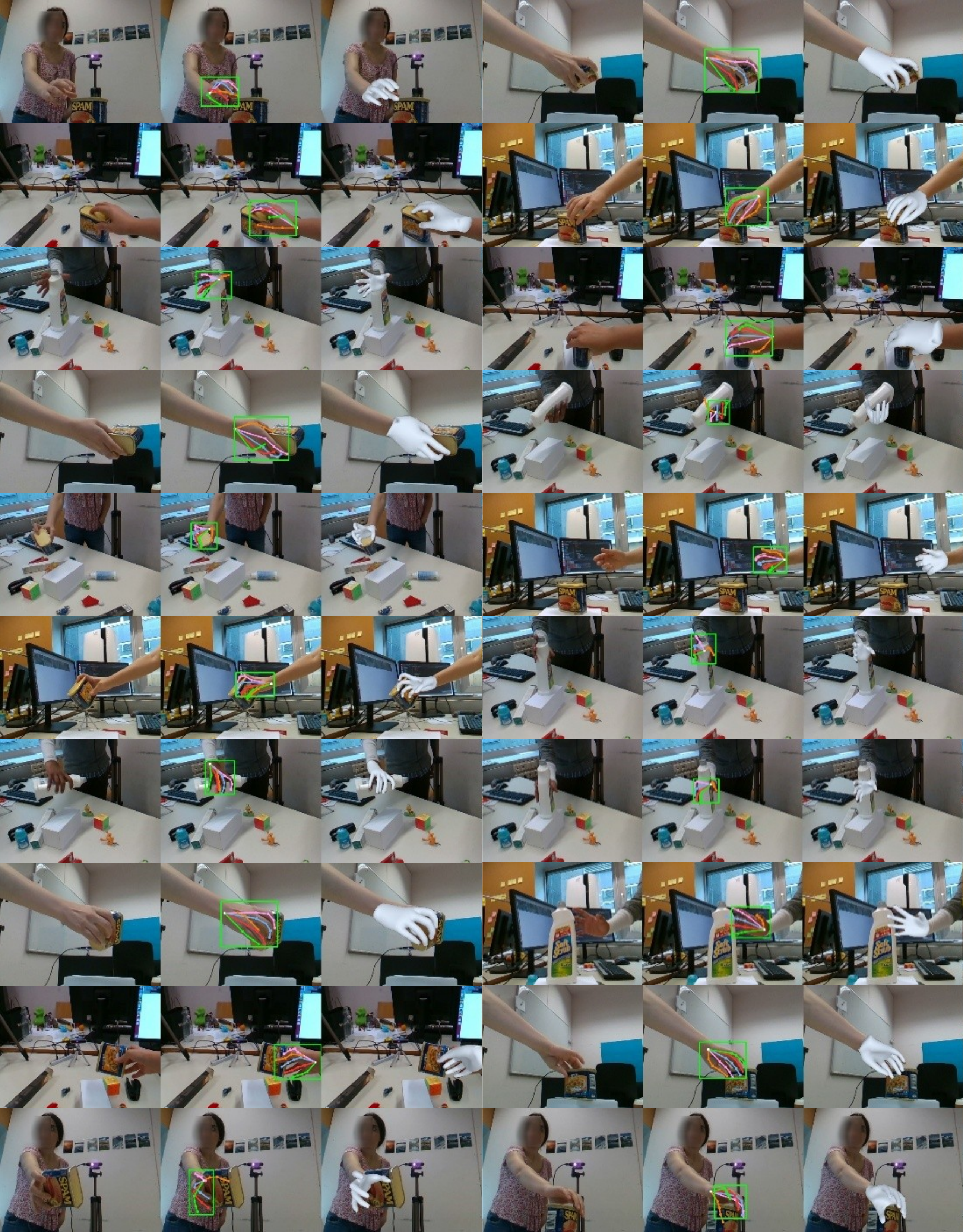}
\caption{
Visualization of HO3Dv3 evaluation set. Samples in a triplet are input, 2D detection and joints, and 3D mesh.
}
\label{fig:supp_ho3d}
\end{center}
\vspace{-0.3cm}
\end{figure*}

\begin{figure*}[t]
\begin{center}
\includegraphics[width=\linewidth]{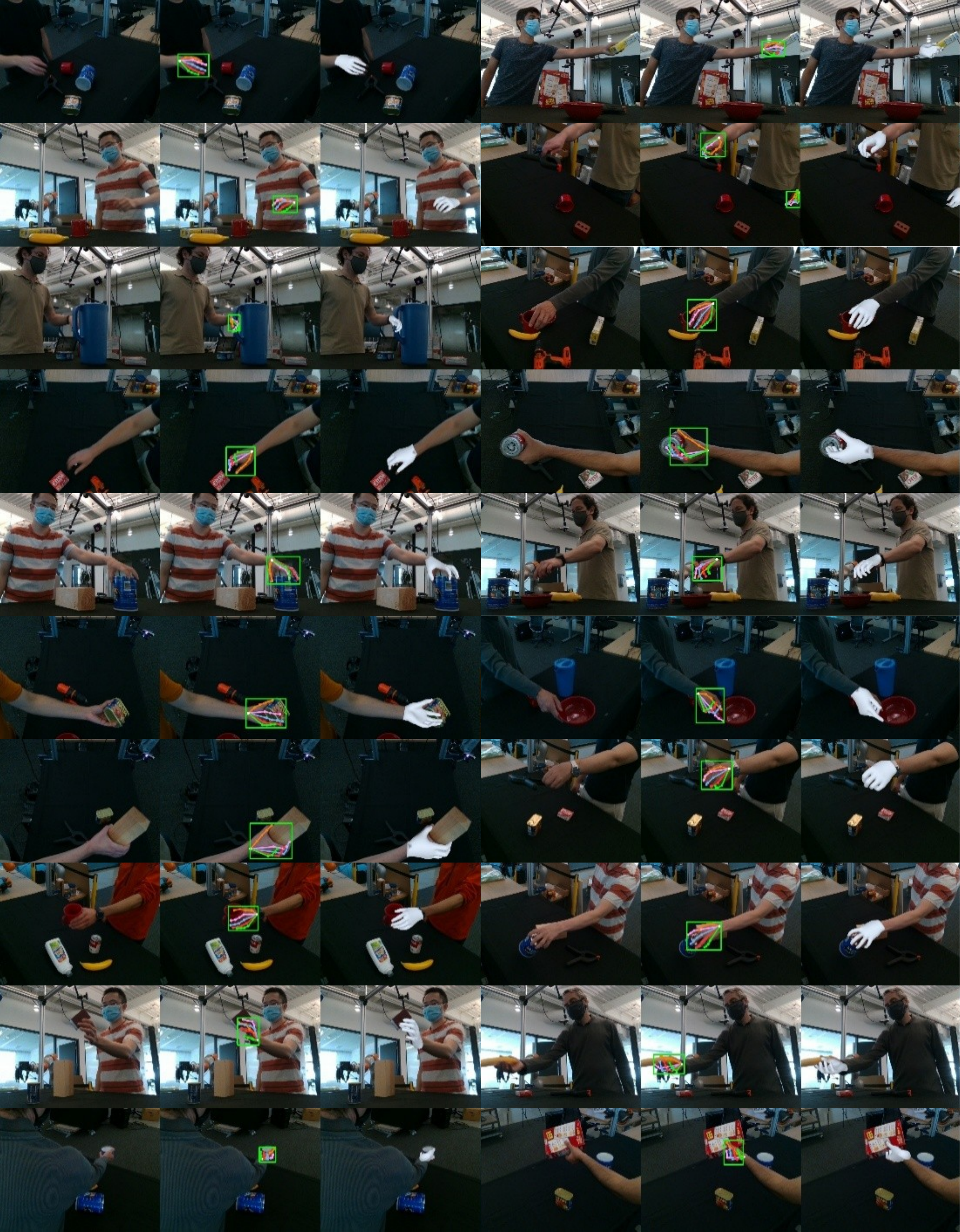}
\caption{
Visualization of DexYCB test set. Samples in a triplet are input, 2D detection and joints, and 3D mesh. 
}
\label{fig:supp_dex}
\end{center}
\vspace{-0.3cm}
\end{figure*}

\begin{figure*}[t]
\begin{center}
\includegraphics[width=\linewidth]{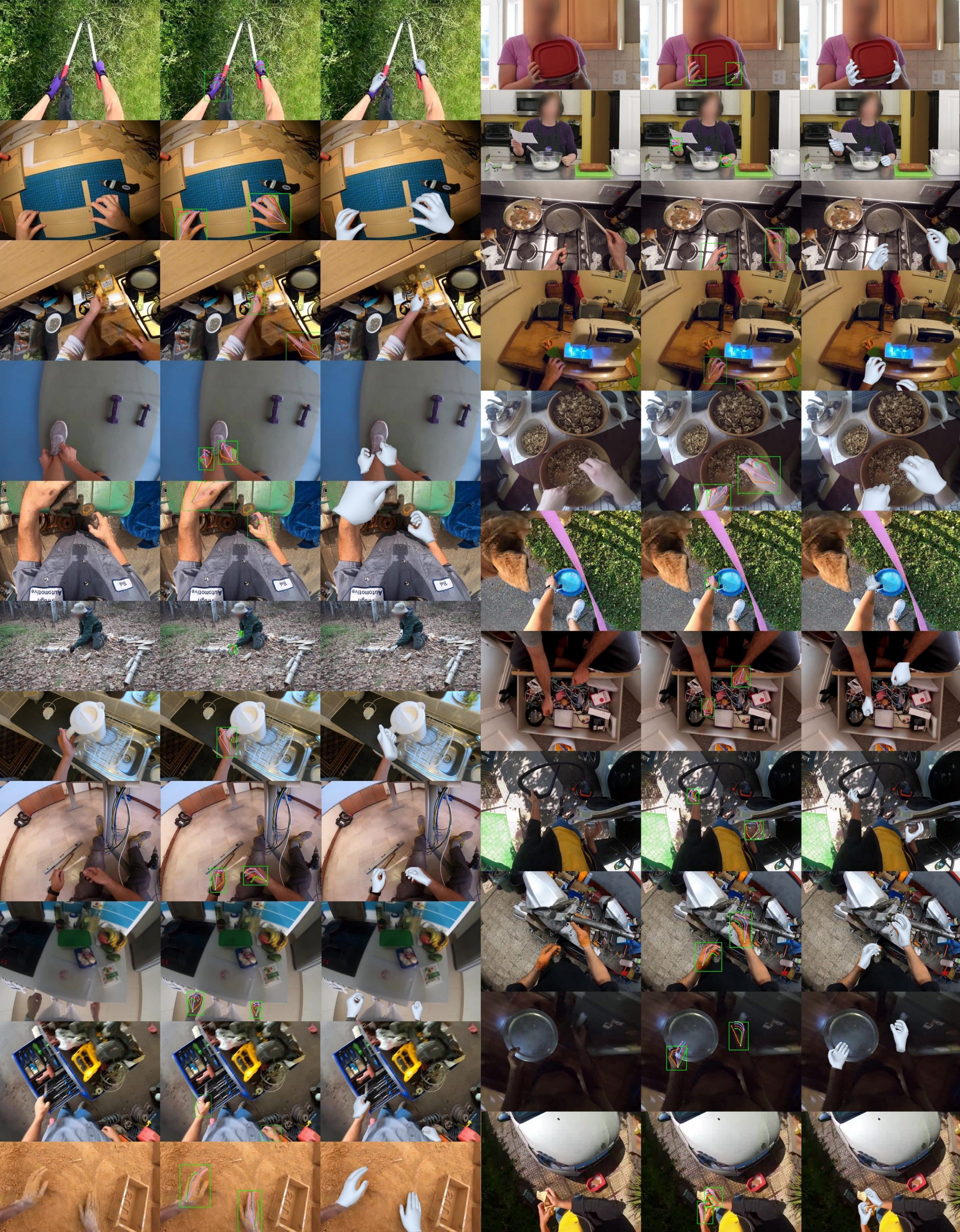}
\caption{
Visualization of HInt test set. Samples in a triplet are input, 2D detection and joints, and 3D mesh. 
}
\label{fig:supp_hit}
\end{center}
\vspace{-0.3cm}
\end{figure*}

\begin{figure*}[t]
\begin{center}
\vspace{-1em}
\includegraphics[width=\linewidth]{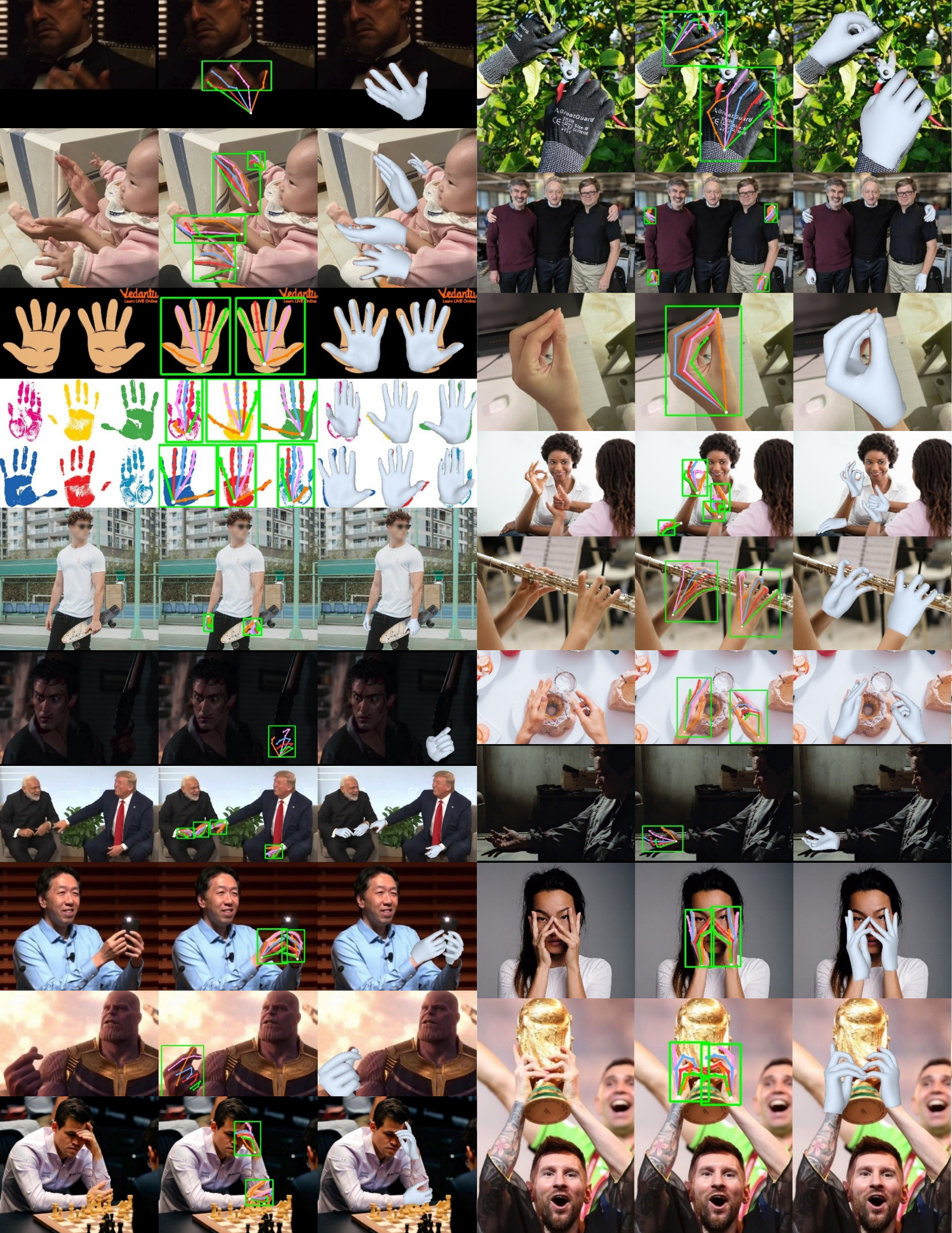}
\caption{
Visualization of practical application. Samples in a triplet are input, 2D detection and joints, and 3D mesh. 
}
\label{fig:supp_pure}
\end{center}
\vspace{-0.3cm}
\end{figure*}

Referring to \autoref{fig:supp_frei}--\ref{fig:supp_hit}, we illustrate samples in our used datasets. As shown, the HandOS can handle various scenarios with hard poses, object occlusion, and \etc. We also demonstrate that our HandOS is capable of real-world applications for difficult textures, shapes, lighting, and styles, as shown in \autoref{fig:supp_pure}. The model for \autoref{fig:supp_pure} is trained with FreiHand \cite{bib:FreiHand}, HInt \cite{bib:Hamer}, CompHand \cite{bib:MobRecon}, COCO \cite{bib:cocow}, OneHand10K \cite{bib:1h10k}. Note that the HandOS exhibits zero-shot generation across styles (\eg, painting, cartoon), benefiting from the open-world representation of Grounding DINO \cite{bib:GD}.


\section{Supplemental Video}
Please refer to our homepage for dynamic results, which demonstrates frame-by-frame processing without employing any temporal strategies.

\section{Limitations and Future Works}
\label{sec:limit}
\paragraph{Geometry prior.}
The HandOS does not incorporate a geometric prior like MANO, meaning that the hand shape is learned entirely from data without relying on any predefined structural knowledge. In our opinion, incorporating an implicit prior (\eg, a variational autoencoder) could accelerate the convergence of HandOS and improve the geometric realism of the predicted hand geometry.


\vspace{-0.3cm}
\paragraph{Pose representation.}
We use keypoints to unify left-right hand representation. Nevertheless, obtaining a rotational pose (\ie $\boldsymbol\theta$ in MANO) is less straightforward and requires an extra inverse kinematics module. 

\vspace{-0.3cm}
\paragraph{Temporal coherence.}
The HandOS is designed for single image processing without considerations for temporal coherence, which may result in jerky outputs when applied to video inference.

\vspace{-0.3cm}
\paragraph{Future works.}
We plan to extend HandOS to provide versatile hand understanding. 
In addition to detection, 2D pose, and 3D mesh, other properties such as segmentation, texture, and object contact are also valuable considerations. 
Furthermore, the HandOS will be utilized to analyze human manipulation skills, contributing to advancements in embodied intelligence.


\end{document}